\documentclass[10pt,twocolumn,letterpaper]{article}

\usepackage[pagenumbers]{cvpr} 

\usepackage[dvipsnames]{xcolor}
\usepackage{multirow}

\usepackage{pifont}
\usepackage{float}

\newcommand{\cmark}{\ding{51}}%
\newcommand{\xmark}{\ding{55}}%
\usepackage[most]{tcolorbox}
\usepackage{colortbl} 
\usepackage{cuted}
\usepackage{listings}
\tcbuselibrary{listingsutf8}
\usepackage{fvextra}

\newcommand{\eb}[1]{$\mathbf{{#1}}$}
\newcommand{\eu}[1]{$\underline{{#1}}$}

\newcommand{\ours}{InterPose\xspace}%
\newcommand{\agent}{HOI-Agent\xspace}%

\definecolor{aliceblue}{rgb}{0.94, 0.97, 1.0}

\usepackage[subtle]{savetrees} 

\definecolor{cvprblue}{rgb}{0.21,0.49,0.74}
\usepackage[pagebackref,breaklinks,colorlinks,citecolor=cvprblue]{hyperref}
\usepackage{soul}

\title{InterPose: Learning to Generate Human-Object Interactions \\from Large-Scale Web Videos}

\author{Yangsong Zhang$^{1}$, 
Abdul Ahad Butt$^1$, Gül Varol$^2$, Ivan Laptev$^1$ \\[2mm]
$^1$Mohamed Bin Zayed University of Artificial Intelligence (MBZUAI) \\
$^2$LIGM, École des Ponts, IP Paris, Univ Gustave Eiffel, CNRS
\\ \url{https://mael-zys.github.io/InterPose/}
}

\begin{document}

\twocolumn[{
\maketitle
\vspace{-0.6cm}
\begin{center}
\begin{minipage}{0.98\textwidth}
    \centering
    \includegraphics[width=\textwidth]{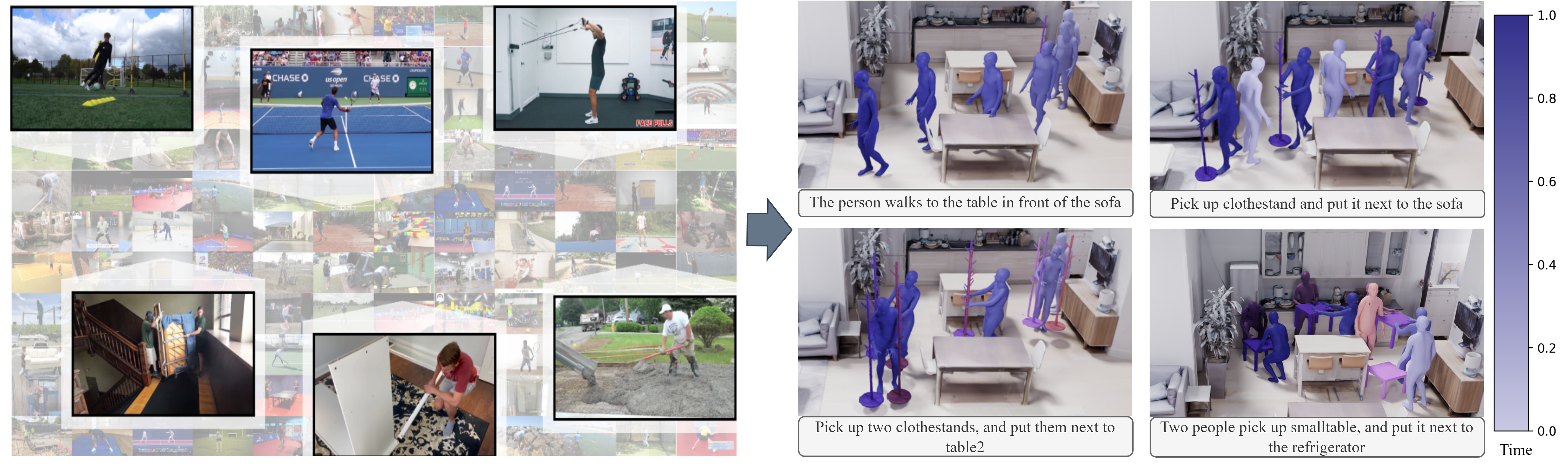}
\end{minipage}
\begin{minipage}{0.49\textwidth}
    \centering
    {\small{(a) Sample videos with Human-Object Interactions (HOI)}}
\end{minipage}
\hfill
\begin{minipage}{0.49\textwidth}
    \centering
    {\small{(b) Zero-shot generation of interactions by our \agent}}
\end{minipage}
\vspace{-0.2cm}
\captionof{figure}{\textbf{Learning from online videos enables generation of complex human motions.
} (a) Our \ours dataset is obtained from videos with varying scenes and activities as well as diverse human-object interactions.
(b) Our \agent deploys \ours for training and enables zero-shot generation of collision-free navigation, human-object interactions and multi-person collaboration in complex 3D scenes.
}
\label{fig:teaser}
\end{center}
}]

\begin{abstract}
Human motion generation has shown great advances powered by recent diffusion models and large-scale motion capture data. 
Most of existing works, however, currently target animation of isolated people in empty scenes.
Meanwhile, synthesizing realistic human–object interactions in complex 3D scenes remains a critical challenge in computer graphics and robotics. 
One obstacle towards generating versatile high-fidelity human-object interactions is the lack of large-scale datasets with diverse object manipulations. 
Indeed, existing motion capture data is typically restricted to single people and manipulations of limited sets of objects.
To address this issue, we propose an automatic motion extraction pipeline
and use it to collect interaction-rich human motions.
Our new dataset \mbox{\ours} contains 73.8K sequences of 3D human motions and corresponding text captions automatically obtained from 45.8K 
videos with human-object interactions.
We perform extensive experiments and demonstrate \ours to bring significant improvements to state-of-the-art methods for human motion generation.
Moreover, using \ours we develop an LLM-based agent enabling zero-shot animation of people interacting with diverse objects and scenes.
\end{abstract}

\section{Introduction}
\label{sec:intro}

Realistic animation of people in complex 3D environments remains a major challenge in gaming, robotics, virtual reality, and embodied AI. 
While games become more realistic, animation of interactions among characters and of object manipulations remains limited. 
In robotics, the deployment of complex manipulation skills is rare and often involves manual teleoperation. 
This motivates the growing interest in automatic generation of realistic interactions aiming to simulate continuous human behavior in 3D scenes.

Large-scale human motion datasets~\cite{mahmood2019amass,guo2022generating,lin2023motion} have substantially advanced motion generation conditioned on past trajectories~\cite{yuan2020dlow,zhang2021we}, action categories~\cite{guo2020action2motion,petrovich21actor} and textual descriptions~\cite{guo2023momask,guo2022generating,petrovich22temos,tevet2022MDM,zhang2023generating}. In contrast, Human–Object Interaction (HOI)~\cite{li2023object,peng2023hoi,li2023controllable,pan2025tokenhsi,gao2024coohoi}, which requires generating coherent human–object motion with physically plausible contacts, remains a more challenging and insufficiently explored problem. Early efforts such as GOAL~\cite{taheri2022goal} and SAGA~\cite{wu2022saga} primarily address small-object grasping, whereas subsequent methods~\cite{diller2024cg,li2023controllable,peng2023hoi,xu2023interdiff} extend to interactions with larger objects. However, these approaches typically depend on the paired human–object training data, exhibit limited generalization, and are often tailored for specific interaction scenarios. This limitation stems from the current HOI datasets~\cite{bhatnagar22behave,li2023object,taheri2020grab,lv2024himonewbenchmarkfullbody}, which remain constrained in both scale and diversity, covering only a narrow spectrum of interaction types.

To address these challenges, in this work we develop a pipeline for automatic collection of large-scale and diverse human motion data from videos. 
We build on advances in 3D human pose estimation~\cite{shin2024wham} and vision–language models (VLMs)~\cite{Qwen2.5-VL} and  extract interaction-rich 3D human motions representing a large variety of daily living activities and sports.
Using this pipeline, we introduce \textbf{\ours}, a large-scale dataset of human motions and corresponding text descriptions, focused on diverse interactions obtained from YouTube videos and open-source video datasets (cf.~Table~\ref{tab:detail_stats}). 
As shown in Table~\ref{tab:data_compare}, \ours is the largest interaction-focused public dataset with human motions. 
As opposed to motion capture datasets taken in controlled settings, \ours also maximizes diversity due to "in the whild" nature of source videos.
We use our dataset to train human animation models MaskedMimic~\cite{tessler2024maskedmimic} and OmniControl~\cite{xie2023omnicontrol}, and demonstrate significant improvements in human motion brought by \ours.

\begin{table}
  \centering
  \resizebox{0.47\textwidth}{!}{
  \begin{tabular}{lrrrrr}
    \toprule
    Dataset & Clip & Frames & Avg Frames & Avg Duration (s) & Hour \\ 
    \midrule
    HD-VILA-100M~\cite{xue2022advancing} & 10,042 & 1.6M & 168.15 & 5.72 & 15.96 \\
    Kinetics-700~\cite{carreira2019short} & 39,464 & 7.5M & 190.21 & 6.40 & 70.20 \\
    Charades~\cite{sigurdsson2016hollywood} & 6,974 & 2.1M & 304.24 & 11.53 & 22.34 \\
    Online videos & 17,334 & 4.4M & 255.59 & 8.36 & 40.24 \\
    \midrule
    Total & 73,814 & 15.7M & 213.34 & 7.25 & 148.74 \\
    \bottomrule
  \end{tabular}
  }
  \vspace{-0.2cm}
  \caption{\textbf{Breakdown of \ours video sources.} \ours is obtained from online YouTube videos and three existing video datasets. 
  }
  \vspace{-0.1cm}
  \label{tab:detail_stats}
\end{table}

\begin{table}
  \centering
  \resizebox{0.47\textwidth}{!}{
  \begin{tabular}{l|c|c|rr|ccc|cc}
    \toprule
    \multirow{2}{*}{Dataset} & \multirow{2}{*}{Year} & \multirow{2}{*}{Source} & \multicolumn{2}{c|}{Scale} & \multicolumn{3}{c|}{Modality} & \multicolumn{2}{c}{Scene} \\ 
     & & & Clip & Hour & Motion & Text & Interaction & Indoor & Outdoor \\
    \midrule
    AMASS~\cite{mahmood2019amass} & 2019 & C & 11,265 & 40 & B,H & \xmark & \xmark & \cmark & \xmark \\
    HumanML3D~\cite{guo2022generating} & 2022 & C & 14,616 & 28.6 & B & \cmark & \xmark & \cmark & \xmark \\
    Motion-X~\cite{lin2023motion} & 2023 & C,V & 81,084 & 144.2 & B,H & \cmark & \xmark & \cmark & \cmark \\
    \midrule
    \midrule
    GRAB~\cite{taheri2020grab} & 2020 & C & 1,334 & 3.8 & B,H & \xmark & \cmark & \cmark & \xmark \\ 
    BEHAVE~\cite{bhatnagar22behave} & 2022 & C & 321 & 4.1 & B & \xmark & \cmark & \cmark & \xmark \\
    OMOMO~\cite{li2023object} & 2023 & C & - & 10.1 & B & \cmark & \cmark & \cmark & \xmark \\
    TRUMANS~\cite{jiang2024scaling} & 2024 & C & - & 15.0 & B,H & \xmark & \cmark & \cmark & \xmark \\
    LINGO~\cite{jiang2024autonomous} & 2024 & C & - & 16.0 & B,H & \cmark & \cmark & \cmark & \xmark \\
    HIMO~\cite{lv2024himonewbenchmarkfullbody} & 2024 & C & 3,376 & 9.4 & B,H & \cmark & \cmark & \cmark & \xmark \\ 
    InterAct-X~\cite{xu2025interact} & 2025 & C & 16,201 & 30.7 & B,H & \cmark & \cmark & \cmark & \xmark \\
    ParaHome~\cite{kim2025parahome} & 2025 & C & 207 & 8.1 & B,H & \cmark & \cmark & \cmark & \xmark \\
    HUMOTO~\cite{lu2025humoto} & 2025 & C & 735 & 2.2 & B,H & \cmark & \cmark & \cmark & \xmark \\ 
    \midrule
    \ours & 2025 & V & 73,814 & 148.7 & B,H & \cmark & \cmark & \cmark & \cmark \\ 
    \bottomrule
  \end{tabular}
  }
  \vspace{-0.2cm}
  \caption{\textbf{Comparison with existing human motion datasets.} Letters C and V mean the data is obtained from motion capture and videos, respectively. B and H present body and hand motion, respectively. Note that, although InterAct-X merges and augments  many existing HOI datasets, \ours provides the largest-scale interaction data.
  }
  \vspace{-0.3cm}
  \label{tab:data_compare}
\end{table}

We leverage the scale and the rich diversity of \ours and demonstrate zero-shot HOI generation using MaskedMimic~\cite{tessler2024maskedmimic}. Experiments on standard HOI benchmarks, including OMOMO~\cite{li2023object} and BEHAVE~\cite{bhatnagar22behave}, demonstrate that our zero-shot approach obtains excellent results and outperforms other in-domain trained methods such as CHOIS~\cite{li2023controllable}.
The benefit of \ours is particularly  prominent for the BEHAVE~\cite{bhatnagar22behave} benchmark, where all models are
assessed in zero-shot settings. 
These results demonstrate that using \ours improves generalization and enables zero-shot generation of human-object interactions for new objects.

To extend interaction generation to complex 3D environments, we further introduce \agent, a zero-shot LLM-based framework. \agent integrates a high-level LLM planner with a low-level motion generator (i.e., MaskedMimic~\cite{tessler2024maskedmimic}). The planner produces structured execution steps, collision-free waypoints, and executable Python code for downstream control, while the \ours-powered motion generator synthesizes human–object interactions in a zero-shot manner.
As illustrated in Figure~\ref{fig:teaser}, our \agent can handle a broad range of interactions within complex 3D scenes.

In summary, our contributions are threefold:
\begin{itemize}[leftmargin=2em]
\item We develop an automatic pipeline for data collection and introduce \ours, a large-scale 3D human motion dataset with diverse interactive activities, facilitating generation of human-object interactions.
\item We use \ours to train state-of-the-art spatial control methods and demonstrate substantial performance improvements enabled by our dataset.
\item We present \agent, a unified zero-shot framework that integrates our human motion model to perform diverse manipulation tasks and generalizes to novel interaction scenarios.
\end{itemize}
Our code, data and models will be made publicly available on the 
project page~\cite{interposewebpage}.

\section{Related work}
\label{sec:related_work}

We briefly review relevant works on controllable motion generation, 3D HOI datasets, as well as HOI generation.

\begin{figure*}
  \centering
   \setlength{\tabcolsep}{8pt}
   \includegraphics[width=0.98\textwidth]{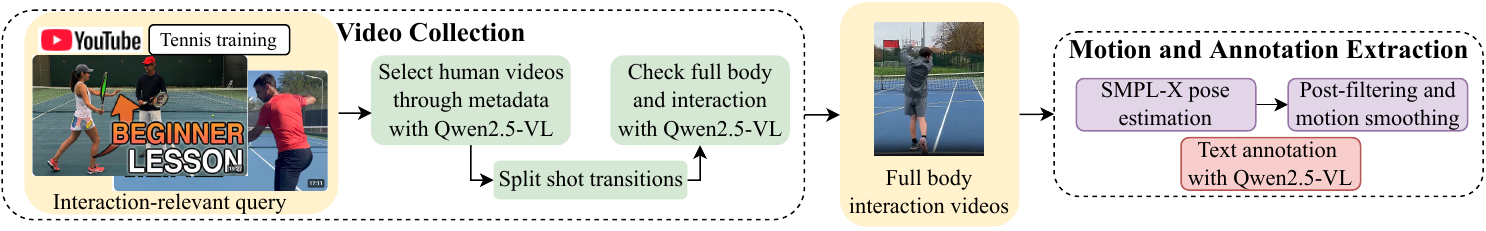}
   \vspace{-0.2cm}
   \caption{\textbf{Overview of data collection for the \ours dataset.} Our framework contains a module for collecting interaction-rich videos (left) and a module for automatic extraction of 3D human motions and corresponding text captions (right).
   }
   \vspace{-0.3cm}
   \label{fig:data_collection}
\end{figure*}

\smallskip
\noindent\textbf{Controllable human motion generation.}
Human motion generation has been explored
under various conditions, including action classes~\cite{guo2020action2motion,petrovich21actor}, textual descriptions~\cite{guo2023momask,tevet2022MDM,zhang2023generating,petrovich22temos}, speech audio~\cite{li2021audio2gestures,zhu2023taming}, music~\cite{li2022danceformer,tseng2023edge}, scene characteristics~\cite{wang2022humanise,zhao2023synthesizing} and objects~\cite{li2023object,li2023controllable}. 
Recent approaches further focus on controllable motion generation methods with text descriptions, including temporal or spatial composition~\cite{TEACH:3DV:2022,SINC:ICCV:2023,petrovich2024multi}, style control~\cite{guo2024generative,jang2022motion} and spatial guidance~\cite{karunratanakul2023guided,xie2023omnicontrol,dai2024motionlcm,tessler2024maskedmimic}. 
To achieve zero-shot object interaction, we sample the contact points and then control the person's hands to move with the object. To this end, the spatial controllability is essential in our zero-shot framework. 

Spatial control models such as OmniControl~\cite{xie2023omnicontrol} and MaskedMimic~\cite{tessler2024maskedmimic} are all trained with human-only datasets~\cite{guo2022generating,mahmood2019amass}, whose data mainly contains 
locomotion and limited interactions. Such models, hence, exhibit difficulties in generating human-object interactions (HOI).
To enable HOI generation, we retrain~\cite{xie2023omnicontrol,tessler2024maskedmimic} using \ours and show improvements.

\smallskip
\noindent\textbf{Human-object interaction generation.}
With the emergence of full-body motion datasets together with hand-object interactions~\cite{taheri2020grab}, models have been developed to synthesize full-body motions leading up to object grasping~\cite{taheri2022goal, wu2022saga}. Subsequent works~\cite{braun2024physically, ghosh2023imos,jiang2024scaling} have expanded this task to human-object interaction for more complex object manipulations. With more recent datasets~\cite{bhatnagar22behave,li2023object}, several works further explore interaction with larger objects~\cite{diller2024cg, li2023controllable,peng2023hoi,xu2023interdiff}. Specifically, CHOIS~\cite{li2023controllable} generates human-object interactions conditioned on sparse 2D object waypoints while HOIFHLI~\cite{wu2024human} further integrates an LLM-based agent to enable the interaction ability in 3D scenes. Previous kinematic-based methods may generate implausible motions such as sliding, floating and human-object penetration. To solve these problems, physics-based methods have recently been explored~\cite{xu2025intermimic,braun2024physically,luo2024grasping,gao2024coohoi,pan2025tokenhsi}.
InterMiMic~\cite{xu2025intermimic} focuses on imitating all the motion capture data in a physics simulator while correcting contact artifacts. 
Prior works highly rely on paired human and object data, and the performance is thus limited by the dataset scale. 
Zero-shot HOI methods
have recently been
explored~\cite{xu2024interdreamer,li2024zerohsi,lou2025zeroshothumanobjectinteractionsynthesis}. 
Specifically, ZeroHSI~\cite{li2024zerohsi} and ZeroHOI~\cite{lou2025zeroshothumanobjectinteractionsynthesis} both leverage a video generation model to first generate human-object videos and then extract the human and object motion through estimation and differentiable rendering. Despite the high interaction quality, the full
process is time-consuming, which takes several hours to obtain one sequence. 
Concurrent work such as HOIGen~\cite{liu2025hoigen} explores HOI video collection in the context of video generation, whereas our focus is on the 3D generation.

Due to data scarcity, existing research, however, mainly focuses on one type of interaction with certain types of objects, and can hardly generalize to some other daily scenarios. Based on \ours, we enable MaskedMimic~\cite{tessler2024maskedmimic} zero-shot HOI ability across various objects. 
Furthermore, we explore an LLM agent framework to perform various HOI tasks such as navigation in 3D scenes and multi-person collaboration while not training on any specific tasks.

\smallskip
\noindent\textbf{Human-object interaction datasets.}
The field of human motion generation has undergone significant development driven by current motion capture datasets. A number of datasets contain 3D human motions in isolation from objects. AMASS~\cite{mahmood2019amass} proposes a large-scale motion capture dataset which enables the development of human motion modeling.  
Motion-X~\cite{lin2023motion} further proposes a larger-scale human-only motion dataset by merging existing datasets and also by extracting motion parameters from videos.  
However, current large-scale datasets mainly include isolated human motions such as walking and sports activities, and contain only a limited set of sequences with interactions.

While there exist human-object datasets with static objects (e.g., scene interaction such as COUCH~\cite{zhang2022couch}, HUMANISE~\cite{wang2022humanise} and GHOST~\cite{milacski2024ghost}), we focus on HOI datasets with dynamic object manipulation.
GRAB~\cite{taheri2020grab} proposes a dynamic HOI dataset where the interaction is mainly focused on small objects. More recent datasets~\cite{bhatnagar22behave,li2023object} have introduced human motion involving manipulation of a wider range of object sizes. 
Prior datasets focus only on single object interaction, to address this, HIMO~\cite{lv2024himonewbenchmarkfullbody} is proposed which specifically focuses on multi-object interaction. Due to the difficulty of collecting large-scale human-object motion, the existing HOI datasets are small-scale. Although InterAct-X~\cite{xu2025interact} merges and augments previous HOI datasets, the total duration (31 hours) is lower compared to AMASS~\cite{mahmood2019amass} (40 hours). The lack of large and diverse datasets hinders the progress of realistic interaction synthesis.
In contrast, as shown in Table~\ref{tab:data_compare}, our proposed dataset, \ours, contains large-scale interactive actions for both indoor and outdoor scenes, spanning 149 hours. We will make \ours and derived models publicly available to foster progress in HOI generation.

\section{Automatic dataset construction}
\label{sec:dataset}

In this paper, we propose a large-scale human motion dataset, \textbf{\ours}, which mainly focuses on interactions.
As illustrated in Figure~\ref{fig:data_collection}, the motion data collection pipeline is composed of 7 main components:
1)~designing keywords to query human-object interaction videos from the web; 
2)~checking video metadata such as thumnail and tags with a VLM before downloading;
3)~pre-processing such as shot splitting 
4)~pre-filtering videos without full-body humans or without interactions;
5)~whole-body 3D pose estimation;
6)~post-processing through smoothing and filtering floating or static humans data;
7)~textual annotation through a VLM.
In Section~\ref{sec:collection}, we present details on the collection of videos with interaction-rich human motion. Automatic human motion estimation and captioning steps are described in Section~\ref{sec:processing}.

\begin{figure*}
  \centering
   \setlength{\tabcolsep}{8pt}
   \includegraphics[width=0.98\textwidth]{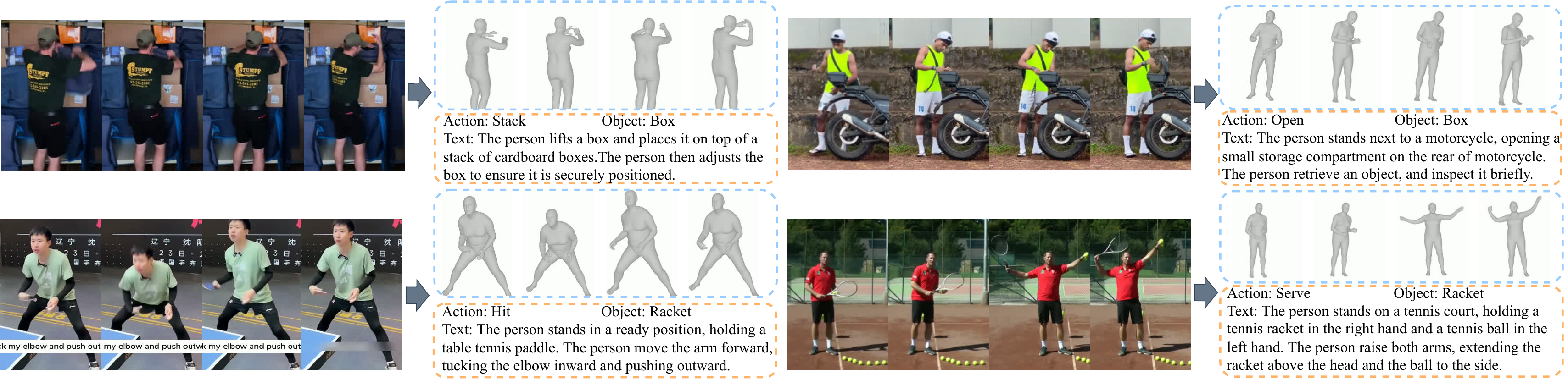}
   \vspace{-0.2cm}
   \caption{\textbf{Examples from \ours.} Our 3D human motion data originates from diverse interaction scenarios including working actions, sports activities, indoor and outdoor scenes. All motion sequences are annotated with action, object label and a detailed textual description.}
   \vspace{-0.3cm}
   \label{fig:example_motion}
\end{figure*}

\subsection{Video collection}
\label{sec:collection}

As detailed in Table~\ref{tab:detail_stats}, we construct \ours by processing 14K raw videos from the indoor action video dataset Charades~\cite{sigurdsson2016hollywood}, 51K from the human action recognition dataset Kinetics-700~\cite{carreira2019short}, and 61K from the high-resolution general video dataset HD-VILA-100M~\cite{xue2022advancing}, yielding 56K motion sequences after filtering and extraction. To further enrich data diversity, we additionally collect 29K online videos and extract 17K motion clips from YouTube. The processing pipeline for existing datasets is identical to that for online videos, except that the query step is omitted. Details of the online video collection are provided in the following paragraphs.

\smallskip
\noindent\textbf{Interaction-relevant query design.} 
Manually searching for human–object interaction videos is both time-consuming and labor-intensive. To address this issue, we leverage a large language model (LLM) to automatically generate a comprehensive set of human interaction queries, enabling systematic coverage of diverse interaction types. Specifically, we ask LLM to give all the queries which may possibly contain full-body and interaction videos such as tennis training, fencing and watering flowers. 
Besides, LLM also provides a variety of queries for similar topics, e.g., for tennis, LLM generates queries such as {\em tennis training}, {\em tennis tutorial}, {\em tennis match} and {\em tennis serve}. 
These queries are integrated into our fully-automated pipeline for searching and downloading online videos.
As shown in Figure~\ref{fig:example_motion}, the collected online videos cover diverse indoor and outdoor scenarios, daily object manipulation actions and sports. More examples can be found in Figure~\ref{fig:more_example_motion} of the Appendix.

\smallskip
\noindent\textbf{Metadata check.} Building upon the query design stage, we incorporate an early-stage filtering mechanism to avoid unnecessary downloads. Specifically, we employ Qwen2.5-VL~\cite{Qwen2.5-VL} to analyze available video metadata, including titles, categories, thumbnails, and ASR-derived speech transcripts, before downloading. Qwen2.5-VL is prompted to verify whether each candidate video predominantly features a full-body human and whether it likely involves interaction with objects or other individuals. This pre-selection significantly reduces the number of irrelevant videos entering subsequent stages.

\smallskip
\noindent\textbf{Shot detection.} Given the downloaded candidate videos, we apply PySceneDetect~\cite{PySceneDetect} to segment videos into coherent temporal clips. When splitting, the frames per second (FPS) of all segments are processed to 30 which is the most common FPS for training motion models. Short segments (i.e., $<$2~sec.) and low resolution ($<$360p.) videos are discarded in this step to maintain motion quality. Besides, long videos are also split into multiple clips of no more than 30 seconds, facilitating later processing and annotation while preserving motion diversity. 

\smallskip
\noindent\textbf{Full-body and interaction verification.} As metadata-based filtering cannot guarantee accuracy, we perform a finer-grained validation on each retained segment. We use a pretrained 2D keypoint detector MediaPipe~\cite{lugaresi2019mediapipe} to infer the keypoints of the most clearly visible human. We then filter the videos where humans are mostly occluded with average confidence less than 0.5 for key joints or the ratio of full-body frames lower than a predefined threshold. In addition, we prompt Qwen2.5-VL~\cite{Qwen2.5-VL} again for all remaining video segments to further confirm that the human consistently engages in physical interactions with objects or other people across most frames. This two-step verification removes residual false positives and ensures the collected clips provide clear, high-quality interaction content.

\begin{table*}
  \centering
  \resizebox{0.99\textwidth}{!}{
  \begin{tabular}{l|l|ccc|ccc|ccc}
    \toprule
    \multirow{2}{*}{Control joints} & \multirow{2}{*}{Training dataset} & \multicolumn{3}{c|}{AMASS~\cite{mahmood2019amass}} & \multicolumn{3}{c|}{OMOMO~\cite{li2023object}} & \multicolumn{3}{c}{BEHAVE\cite{bhatnagar22behave}} \\ 
     & &  Succ 0.5 $\uparrow$ &  Succ 0.2 $\uparrow$ & MPJPE $\downarrow$ &  Succ 0.5 $\uparrow$ &  Succ 0.2 $\uparrow$ & MPJPE $\downarrow$ &  Succ 0.5 $\uparrow$ &  Succ 0.2 $\uparrow$ & MPJPE $\downarrow$ \\
    \midrule
    \multirow{4}{*}{Pelvis} & AMASS~\cite{mahmood2019amass} & \eb{0.9657} & \eu{0.8673} & \eb{46.1} & \eu{0.9834} & 0.9087 & \eu{49.0} & \eu{0.9932} & 0.9444 & \eu{49.9} \\
     & OMOMO~\cite{li2023object} & 0.8146 & 0.5748 & 150.4 & 0.9585 & 0.7759 & 70.0 & 0.9824 & 0.8035 & 74.8 \\
    & \ours & 0.9228 & 0.8021 & 97.6 & \eb{0.9979} & \eu{0.9191} & 59.0 & 0.9892 & \eu{0.9593} & 55.0 \\
     & \ours + AMASS~\cite{mahmood2019amass} & \eu{0.9636} & \eb{0.9135} & \eu{55.4} & \eb{0.9979} & \eb{0.9855} & \eb{41.2} & \eb{0.9959} & \eb{0.9905} & \eb{37.1} \\
     \midrule
    \multirow{4}{*}{Hands} & AMASS~\cite{mahmood2019amass} & \eu{0.9451} & 0.7841 & \eu{72.7} & 0.8382 & 0.5290 & 133.9 & 0.9837 & 0.7724 & 89.8 \\
     & OMOMO~\cite{li2023object} & 0.7145 & 0.2757 & 207.5 & 0.9170 & 0.6017 & 116.5 & 0.9688 & 0.6125 & 121.8 \\
     & \ours & 0.8831 & \eu{0.8010} & 131.1 & \eu{0.9191} & \eu{0.8071} & \eu{79.2} & \eu{0.9919} & \eu{0.9607} & \eu{65.6} \\
     & \ours + AMASS~\cite{mahmood2019amass} & \eb{0.9521} & \eb{0.8717} & \eb{71.5} & \eb{0.9751} & \eb{0.9004} & \eb{60.9} & \eb{0.9973} & \eb{0.9837} & \eb{52.5} \\
    \bottomrule
  \end{tabular}
  }
  \vspace{-0.2cm}
  \caption{\textbf{Evaluation of human motion controllability using physics-based  MaskedMimic~\cite{tessler2024maskedmimic} generator.} We train MaskedMimic~\cite{tessler2024maskedmimic} with different datasets and then evaluate all the models with different control settings on the human-only dataset AMASS~\cite{mahmood2019amass}, and object interaction datasets (OMOMO~\cite{li2023object} and BEHAVE~\cite{bhatnagar22behave}). The best results are in bold, and the second best results are underlined.
  }
  \label{tab:spatial_control_mask}
\end{table*}
\begin{table*}
  \setlength{\tabcolsep}{2pt}
  \centering
  \resizebox{0.99\textwidth}{!}{
  \begin{tabular}{l|l|cccc|cccc|cccc}
    \toprule
    \multirow{2}{*}{Control joints} & \multirow{2}{*}{Training dataset} & \multicolumn{4}{c|}{HumanML3D~\cite{guo2022generating}} & \multicolumn{4}{c|}{OMOMO~\cite{li2023object}} & \multicolumn{4}{c}{BEHAVE\cite{bhatnagar22behave}} \\ 
     & & FS $\downarrow$ & Traj err 0.5 $\downarrow$ & Traj err 0.2 $\downarrow$ & MPJPE $\downarrow$ & FS $\downarrow$ & Traj err 0.5 $\downarrow$ & Traj err 0.2 $\downarrow$ & MPJPE $\downarrow$ & FS $\downarrow$ & Traj err 0.5 $\downarrow$ & Traj err 0.2 $\downarrow$ & MPJPE $\downarrow$ \\
    \midrule
    \multirow{4}{*}{Pelvis} & HumanML3D~\cite{guo2022generating} & \eu{0.0585} & \eu{0.0957} & 0.3467 & 0.0708 & \eu{0.0945} & 0.0604 & 0.5208 & 0.0836 & 0.0497 & 0.0179 & 0.3259 & 0.0668 \\
     & OMOMO~\cite{li2023object} & 0.0830 & 0.3125 & 0.5283 & 0.1786 & 0.1175 & 0.0646 & 0.4375 & 0.0719 & \eu{0.0334} & \eu{0.0045} & 0.2277 & 0.0492 \\
     & \ours & \eb{0.0539} & \eu{0.0957} & \eu{0.3066} & \eu{0.0708} & \eb{0.0779} & \eu{0.0208} & \eu{0.2292} & \eu{0.0515} & \eb{0.0320} & \eb{0} & \eu{0.0982} & \eu{0.0342} \\
     & \ours + HumanML3D~\cite{guo2022generating} & 0.0656 & \eb{0.0508} & \eb{0.2207} & \eb{0.0489} & 0.1185 & \eb{0.0042} & \eb{0.1167} & \eb{0.0327} & 0.0488 & \eb{0} & \eb{0.0402} & \eb{0.0232} \\
     \midrule
    \multirow{4}{*}{Hands} & HumanML3D~\cite{guo2022generating} & 0.0681 & \eu{0.1377} & \eu{0.5645} & \eu{0.0849} & 0.0960 & 0.2313 & 0.8646 & 0.1275 & 0.0878 & 0.0670 & 0.6741 & 0.0911 \\
     & OMOMO~\cite{li2023object} & 0.1066 & 0.4736 & 0.8936 & 0.2016 & 0.1408 & 0.1187 & 0.8771 & 0.1410 & 0.0768 & 0.1696 & 0.7232 & 0.1223 \\
     & \ours & \eb{0.0436} & 0.1699 & 0.6123 & 0.0950 & \eb{0.0596} & \eu{0.0979} & \eu{0.6312} & \eb{0.0059} & \eb{0.0275} & \eu{0.0580} & \eu{0.3839} & \eu{0.0591} \\
    & \ours + HumanML3D~\cite{guo2022generating} & \eu{0.0524} & \eb{0.0605} & \eb{0.3320} & \eb{0.0555} & \eu{0.0741} & \eb{0.0354} & \eb{0.3396} & \eu{0.0461} & \eu{0.0639} & \eb{0.0045} & \eb{0.2098} & \eb{0.0320} \\
    \bottomrule
  \end{tabular}
  }
  \vspace{-0.2cm}
  \caption{\textbf{Evaluation of human motion controllability using diffusion-based OmniControl~\cite{xie2023omnicontrol} generator.} 
  Similar to Table~\ref{tab:spatial_control_mask}, we train OmniControl~\cite{xie2023omnicontrol} with different datasets and then evaluate on HumanML3D~\cite{guo2022generating}, OMOMO~\cite{li2023object} and BEHAVE~\cite{bhatnagar22behave}.
  }
  \label{tab:spatial_control_omni}
\end{table*}

\subsection{3D motion and annotation extraction}
\label{sec:processing}

\smallskip
\noindent\textbf{Pose estimation.} Our objective is to extract plausible body pose parameters (i.e.,~\mbox{SMPL-X}~\cite{pavlakos2019expressive}) with both body and hand motions, and to obtain a text description for each interaction.
In terms of body pose, we leverage WHAM~\cite{shin2024wham} to jointly estimate the local human pose and the global transformations in world coordinates. As WHAM only estimates SMPL~\cite{loper2015smpl} parameters, we further apply HaMeR~\cite{pavlakos2024reconstructing} to extract the detailed finger motion. To enhance the consistency between body and hands predictions, we input the same ViTPose++~\cite{xu2023vitpose++} 2D pose predictions for WHAM~\cite{shin2024wham} and HaMeR~\cite{pavlakos2024reconstructing} pipelines. An additional optimization is applied to merge body and hand parameters to avoid twisted wrists.
Note that we extract only the human motion, and not the object, mainly due to the limited performance of current 3D object reconstruction methods. We show that this is already a crucial step for obtaining large-scale human motion data with interaction priors. Future work can revisit complementing the data with object motion.

\smallskip
\noindent\textbf{Post-filtering and motion smoothing.}
Human bounding boxes that are relatively small within a video make it difficult for pose estimation and VLM to capture accurate information. We therefore discard detected people with bounding box area below a threshold of 5,000p. 
We also detect and discard motions with abrupt translations or orientations, e.g., the motion may jump from one person to another because of the detection errors. 
Furthermore, the human actions are manually slowed down or even paused in some videos such as sports tutorial videos.
We discard static data by setting a threshold on joint velocities.

Finally, due to object occlusions, motion blur, or camera movement, the initial SMPL-X~\cite{pavlakos2019expressive} results may exhibit jitter issues. 
To tackle jitter problems, we first detect frames with severe discontinuities in human motion and apply Slerp to interpolate motion trajectories.
We then apply another smoothing step to all frames using a small window size to remove the small-scale jitter while preserving motion details. 

\smallskip
\noindent\textbf{Text annotation.} In terms of textual motion descriptions, we carefully prompt a VLM, Qwen2.5-VL~\cite{Qwen2.5-VL}, to automatically annotate each human 
in the candidate videos. As \ours mainly focuses on interactive actions,
we ask the VLM
to give the whole-body action, a detailed text about the contact body part and the corresponding objects. In the multi-person video cases, to ensure the correspondence between the human and each text, we
crop the video using
the bounding box given from previous ViTPose++~\cite{xu2023vitpose++} predictions. 
As illustrated in Figure~\ref{fig:example_motion}, in addition to textual motion descriptions, the VLM is also used to annotate the human action and the interacted object. All the detailed prompts can be found in Section~\ref{sec:supp_annotation_prompt} of the Appendix.

\section{Experiments}
\label{sec:experiments}

\begin{table*}
  \centering
  \resizebox{0.99\textwidth}{!}{
  \begin{tabular}{l|c|cccc|cccc|cccc}
    \toprule
    \multirow{2}{*}{Control joints} & \multirow{2}{*}{Filtering} & \multicolumn{4}{c|}{HumanML3D~\cite{guo2022generating}} & \multicolumn{4}{c|}{OMOMO~\cite{li2023object}} & \multicolumn{4}{c}{BEHAVE\cite{bhatnagar22behave}} \\ 
     & & FS $\downarrow$ & Traj err 0.5 $\downarrow$ & Traj err 0.2 $\downarrow$ & MPJPE $\downarrow$ & FS $\downarrow$ & Traj err 0.5 $\downarrow$ & Traj err 0.2 $\downarrow$ & MPJPE $\downarrow$ & FS $\downarrow$ & Traj err 0.5 $\downarrow$ & Traj err 0.2 $\downarrow$ & MPJPE $\downarrow$ \\
    \midrule
    \multirow{2}{*}{Pelvis} & \xmark & 0.0713 & 0.0527 & 0.2441 & 0.0522 & 0.1215 & \eb{0} & 0.1208 & 0.0346 & 0.0622 & \eb{0} & 0.0622 & 0.0268 \\
     & \cmark & \eb{0.0656} & \eb{0.0508} & \eb{0.2207} & \eb{0.0489} & \eb{0.1185} & 0.0042 & \eb{0.1167} & \eb{0.0327} & \eb{0.0488} & \eb{0} & \eb{0.0402} & \eb{0.0232} \\
     \midrule
    \multirow{2}{*}{Hands} & \xmark & 0.0683 & 0.0703 & 0.3584 & 0.0590 & 0.0791 & \eb{0.0333} & \eb{0.3292} & 0.0470 & 0.0684 & 0.0089 & 0.2188 & 0.0337 \\
     & \cmark & \eb{0.0524} & \eb{0.0605} & \eb{0.3322} & \eb{0.0555} & \eb{0.0741} & 0.0354 & 0.3396 & \eb{0.0461} & \eb{0.0639} & \eb{0.0045} & \eb{0.2098} & \eb{0.0320} \\
    \bottomrule
  \end{tabular}
  }
  \vspace{-0.2cm}
  \caption{\textbf{Ablation on data filtering with 
  	OmniControl~\cite{xie2023omnicontrol}.} We train OmniControl on the combination of HumanML3D~\cite{guo2022generating} and \ours with/without filtering. All the models are evaluated on HumanML3D~\cite{guo2022generating}, OMOMO~\cite{li2023object} and BEHAVE~\cite{bhatnagar22behave} test sets with pelvis or hands control.
  }
  \label{tab:ablation_filtering}
\end{table*}

\begin{figure*} 
 \centering
  \begin{subfigure}[b]{0.33\textwidth}
    \includegraphics[width=\linewidth]{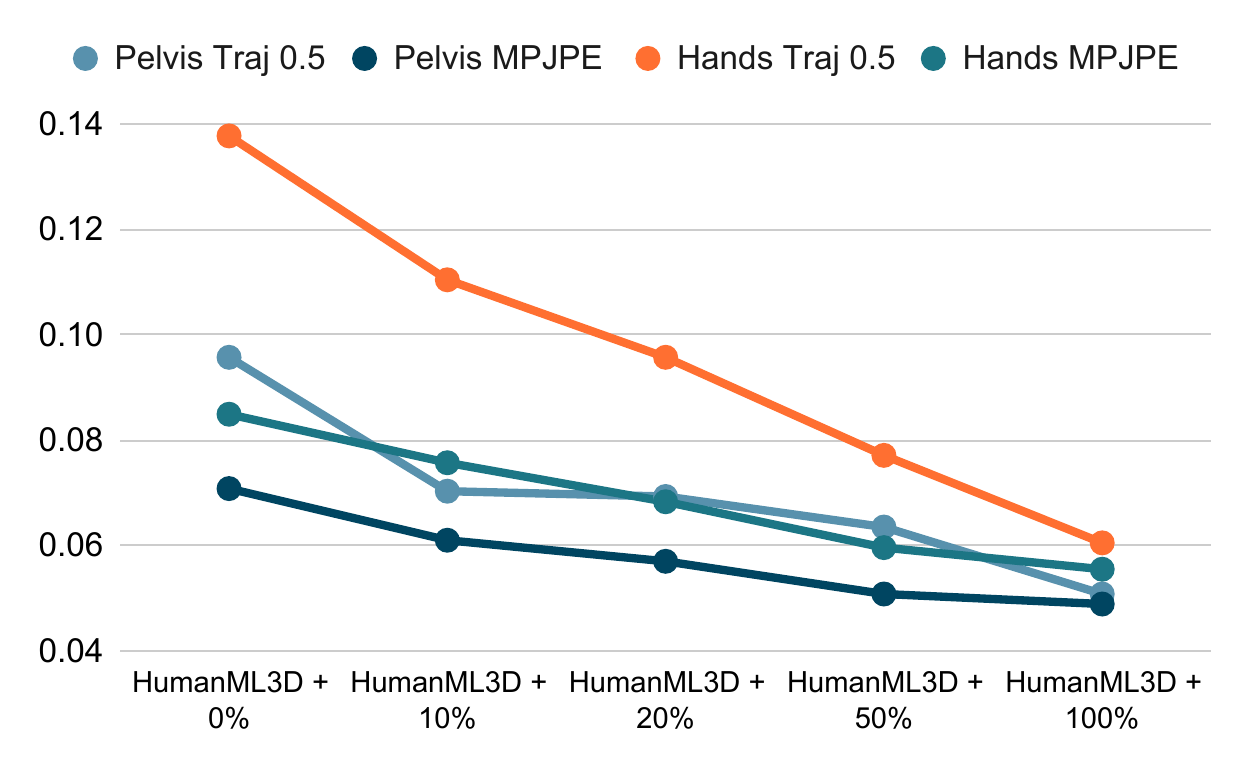}
    \caption{Evaluation on HumanML3D~\cite{guo2022generating} test set}
    \label{fig:data_ratio_human}
  \end{subfigure}
  \hfill 
  \begin{subfigure}[b]{0.33\textwidth}
    \includegraphics[width=\linewidth]{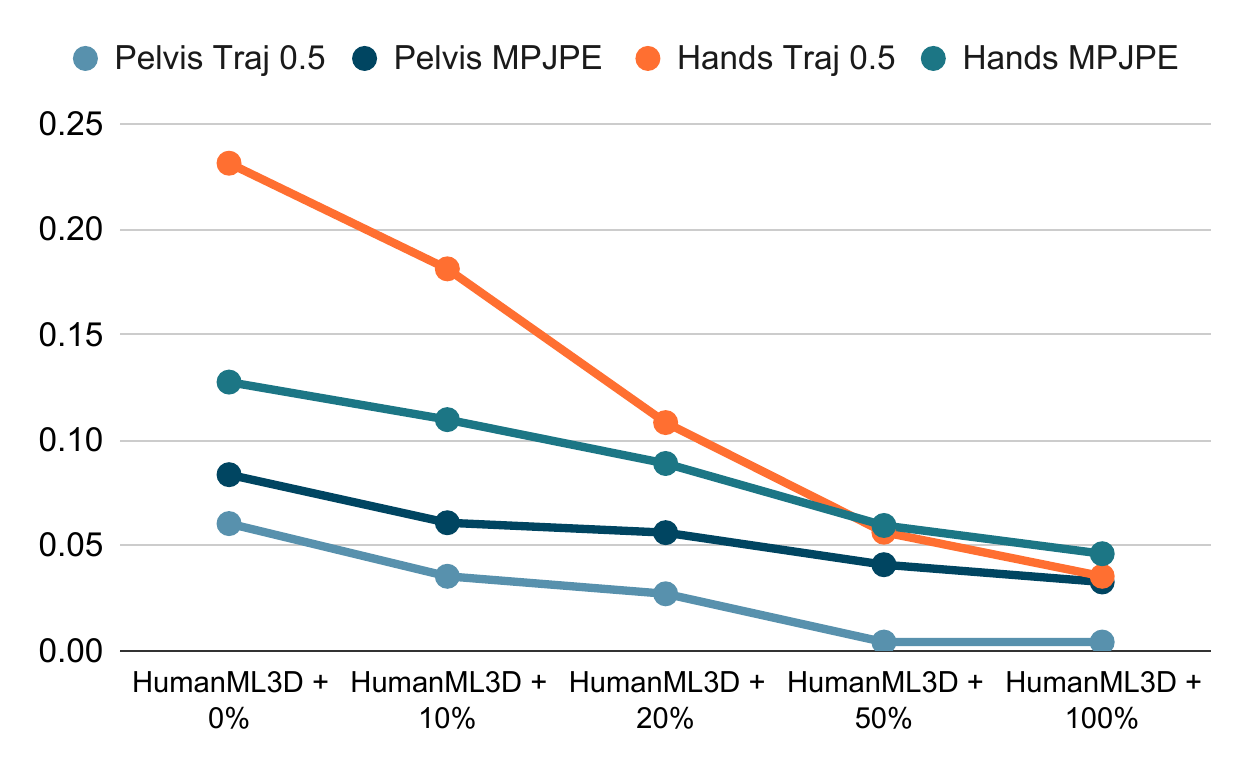}
    \caption{Evaluation on OMOMO~\cite{li2023object} test set}
    \label{fig:data_ratio_omomo}
  \end{subfigure}
  \hfill 
  \begin{subfigure}[b]{0.33\textwidth}
    \includegraphics[width=\linewidth]{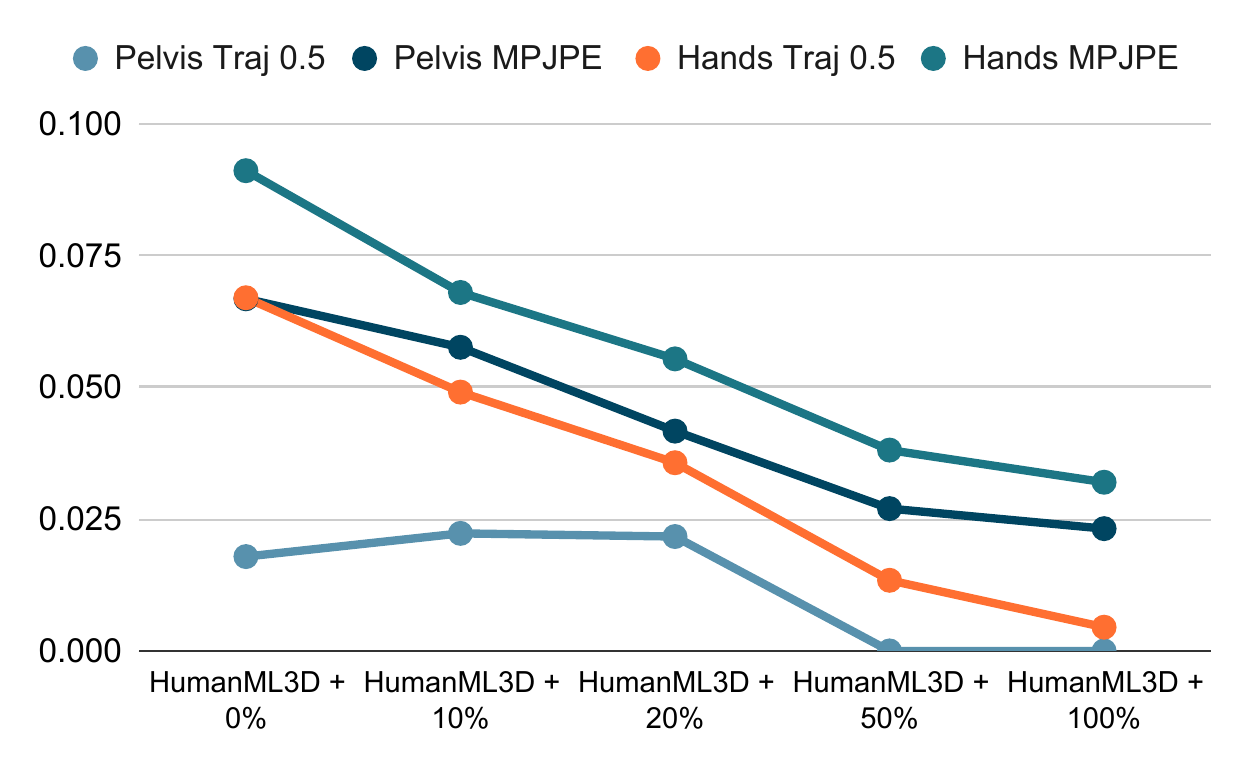}
    \caption{Evaluation on BEHAVE~\cite{bhatnagar22behave} test set}
    \label{fig:data_ratio_behave}
  \end{subfigure}
  \vspace{-0.5cm}
  \caption{\textbf{Impact of dataset size.} We train OmniControl~\cite{xie2023omnicontrol} on HumanML3D~\cite{guo2022generating} dataset and subsets of \ours of different sizes. The subsets are composed of 0\%, 10\%, 20\%, 50\%, and 100\% of \ours training set. All the models are evaluated on HumanML3D~\cite{guo2022generating}, OMOMO~\cite{li2023object} and BEHAVE~\cite{bhatnagar22behave} test sets. We report Traj 0.5 and MPJPE on pelvis control and hands control setting for all the models.}
  \label{fig:dataset}
\end{figure*}

\begin{table}
  \centering
  \resizebox{0.47\textwidth}{!}{
  \begin{tabular}{l|c|ccc|ccc}
    \toprule
    \multirow{2}{*}{Eval joints} & \multirow{2}{*}{Training joints} & \multicolumn{3}{c|}{Full trajectory}  & \multicolumn{3}{c}{Avg}\\ 
     & & Traj err 0.5 $\downarrow$ & Traj err 0.2 $\downarrow$ & MPJPE $\downarrow$  & Traj err 0.5 $\downarrow$ & Traj err 0.2 $\downarrow$ & MPJPE $\downarrow$ \\
    \midrule
    \multirow{2}{*}{Pelvis} & One~\cite{xie2023omnicontrol} & \eb{0.0605} & \eb{0.2637} & \eb{0.0564} & \eb{0.0404} & \eb{0.1623} & \eb{0.0367} \\
    & Random & 0.0957 & 0.3467 & 0.0708 & 0.0633 & 0.2182 & 0.0499 \\
     \midrule
    \multirow{2}{*}{Hands} & One~\cite{xie2023omnicontrol} & 0.5195 & 0.9014 & 0.2161 & 0.2482 & 0.4730 & 0.1102 \\
     & Random & \eb{0.1377} & \eb{0.5645} & \eb{0.0849} & \eb{0.0686} & \eb{0.2639} & \eb{0.0423} \\
    \bottomrule
  \end{tabular}
  }
  \vspace{-0.2cm}
  \caption{\textbf{Ablation for OmniControl~\cite{xie2023omnicontrol} training setting on the HumanML3D~\cite{guo2022generating} dataset.} We train OmniControl~\cite{xie2023omnicontrol} with different number of joints and then evaluate with 2 settings on HumanML3D~\cite{guo2022generating}. Avg means that we evaluate with 5 sparsity levels in the controlling signal, including 1, 2, 5, 25\% density, and 100\% density (full trajectory) and then report the average performance.
  }
  \label{tab:ablation_omnicontrol}
\end{table}

In this section we present advances in human motion generation brought by~\ours. In particular, we deploy \ours to train two state-of-the-art methods, namely OmniControl~\cite{xie2023omnicontrol} and MaskedMimic~\cite{tessler2024maskedmimic}, and demonstrate improvements on standard benchmarks for controlled motion generation in Section~\ref{subsec:exp_control}.
\ours analysis and ablation experiments are presented in Section~\ref{subsec:ablation}. 
In Section~\ref{subsec:zero_shot} we evaluate \ours benefits on the task of generating Human-Object Interactions (HOI). Finally, we present a HOI-Agent and demonstrate its zero-shot generalization to new objects and tasks in Section~\ref{subsec:agent}.

\subsection{Controllable motion generation}
\label{subsec:exp_control}

\noindent\textbf{Implementation details.} To validate the advantage of \ours, we train spatial control models (i.e., OmniControl~\cite{xie2023omnicontrol} and MaskedMimic~\cite{tessler2024maskedmimic}) on different datasets: general human motion datasets (AMASS~\cite{mahmood2019amass} and HumanML3D~\cite{guo2022generating}) and interaction motion datasets (OMOMO~\cite{li2023object} and \ours).
Since interactive tasks often involve multiple control joints, we modify the original training procedure of OmniControl~\cite{xie2023omnicontrol}, which samples only one joint per iteration, by allowing a random number of joints to be controlled. This adjustment better reflects real application scenarios. An ablation study is provided in Section~\ref{subsec:ablation}. Otherwise, the training settings follow those of OmniControl~\cite{xie2023omnicontrol} and MaskedMimic~\cite{tessler2024maskedmimic}.
To evaluate spatial controllability, we condition on full joint trajectories for motion capture datasets AMASS~\cite{mahmood2019amass}, HumanML3D~\cite{guo2022generating}, and OMOMO~\cite{li2023object}, as most interactions require consistent contact along the trajectory. Additionally, we assess zero-shot generalization on BEHAVE~\cite{bhatnagar22behave} to examine performance under unseen control distributions.

\noindent\textbf{Metrics.}
Following MaskedMimic~\cite{tessler2024maskedmimic} and OmniControl~\cite{xie2023omnicontrol}, the spatial controllability is evaluated by the success rate (Succ), failure rate (Traj err) and mean per-joint position error (MPJPE) to calculate the overall accuracy and position error on the control joints. For example, Succ 0.5 denotes the proportion of generated samples whose maximum keyframe position error is below 0.5m. In addition, we report the foot sliding (FS) metric for OmniControl~\cite{xie2023omnicontrol} evaluation to assess motion realism.

\noindent\textbf{Results.}
The performance of MaskedMimic~\cite{tessler2024maskedmimic} and OmniControl~\cite{xie2023omnicontrol} trained on different datasets is presented in Tables~\ref{tab:spatial_control_mask}-\ref{tab:spatial_control_omni} respectively. 
For MaskedMimic~\cite{tessler2024maskedmimic}, training on AMASS~\cite{mahmood2019amass} yields strong performance under pelvis control due to the diverse locomotion of AMASS~\cite{mahmood2019amass}, which helps the model follow pelvis trajectories. 
However, controlling two hands is more challenging and closely tied to interaction, leading to a significant performance drop. In this setting, training on \ours substantially outperforms AMASS~\cite{mahmood2019amass}, especially on the test sets of interaction datasets OMOMO~\cite{li2023object} and BEHAVE~\cite{bhatnagar22behave}.
Since \ours contains no motions from these datasets, the improved generalization highlights the quality and diversity of its interactions. Combining AMASS and \ours achieves the best overall performance.
Similarly, for OmniControl~\cite{xie2023omnicontrol}, training with \ours notably enhances hand controllability.
As MaskedMimic~\cite{tessler2024maskedmimic} and OmniControl~\cite{xie2023omnicontrol} represent fundamentally different paradigms, their consistent improvements demonstrate the robustness of the \ours dataset.

\subsection{Ablations}
\label{subsec:ablation}

\noindent\textbf{Impact of post-filtering.} We evaluate the effect of post-filtering described in Section~\ref{sec:processing} by comparing the spatial controllability of OmniControl~\cite{xie2023omnicontrol} trained on the combined HumanML3D and \ours datasets, with and without post-filtering, as shown in Table~\ref{tab:ablation_filtering}. Model with post-filtering achieves better overall performance. Although the unfiltered model performs slightly worse, it still follows the control trajectory reasonably well from a spatial controllability perspective. However, the lack of filtering introduces abnormal artifacts such as floating and abrupt rotations, leading to significantly increased foot sliding. These results highlight the importance of post-filtering for improving motion realism.

\smallskip
\noindent\textbf{Impact of dataset size.} We investigate how dataset size influences performance to assess whether larger scale interaction data further enhances spatial control. OmniControl~\cite{xie2023omnicontrol} is trained on HumanML3D~\cite{guo2022generating} combined with varying proportions of \ours data (0$\%$, 10$\%$, 20$\%$, 50$\%$, and 100$\%$). As shown in Figure~\ref{fig:dataset}, spatial controllability is evaluated on HumanML3D~\cite{guo2022generating}, OMOMO~\cite{li2023object}, and BEHAVE~\cite{bhatnagar22behave} test sets. Performance improves consistently as the amount of training data increases across all control settings and benchmarks, suggesting that collecting additional motion data could yield further gains. Detailed quantitative results are provided in Table~\ref{tab:ablation_data_ratio} of the Appendix.

\smallskip
\noindent\textbf{OmniControl training setting.}
We further investigate the impact of training configurations in OmniControl~\cite{xie2023omnicontrol}. The original model is trained by randomly selecting a single control joint at each iteration. In contrast, we retrain the model by sampling a random number of control joints per iteration and compare its performance with the original setting. 
Evaluation is conducted under five levels of control signal sparsity, and we report both the full-trajectory results and average performance across sparsity levels. Note that, the Traj 0.5 and MPJPE for ”Avg“ column of  OmniControl~\cite{xie2023omnicontrol} pelvis control are from their published results. Otherwise, we report our reproduced numbers. As shown in Table~\ref{tab:ablation_omnicontrol}, while our strategy shows a slight decrease in pelvis control accuracy, it consistently yields substantial improvements in hand control performance.

\begin{table*}
  \setlength{\tabcolsep}{8pt}
  \centering
  \resizebox{0.8\textwidth}{!}{
  \begin{tabular}{l|l|cc|ccccc}
    \toprule
    \multirow{2}{*}{Model} & \multirow{2}{*}{Training datasets} & \multicolumn{2}{c|}{Human motion} & \multicolumn{5}{c}{Interaction} \\ 
   & & $\text{H}_{\text{feet}}$ $\downarrow$ & FS $\downarrow$ & $\text{C}_{\text{prec}}$ $\uparrow$ & $\text{C}_{\text{rec}}$ $\uparrow$ & $\text{C}_{\text{f1}}$ $\uparrow$ & $\text{C}_{\text{percent}}$  & $\text{P}_{\text{hand}}$ $\downarrow$ \\
    \midrule
    CHOIS~\cite{li2023controllable} & OMOMO~\cite{li2023object} & 4.47 & \eu{0.3435} & \eb{0.7982} & 0.6477 & 0.6773 & 0.5602 & 0.6135 \\
     \midrule
    \multirow{4}{*}{MaskedMimic~\cite{tessler2024maskedmimic}} & AMASS~\cite{mahmood2019amass} & \eu{0.15} & \eb{0.3317} & 0.7882 & 0.6197 & 0.6484 & 0.6225 & 0.6565 \\
     & OMOMO~\cite{li2023object} & 0.76 & 0.4752 & 0.7798 & 0.6032 & 0.6335 & 0.5978 & \eb{0.5876} \\
     & \ours & 0.47 & 0.4249 & 0.7890 & \eu{0.7185} & \eu{0.7172} & \eu{0.7134} & \eu{0.6131} \\
    & \ours + AMASS~\cite{mahmood2019amass} & \eb{-0.11} & 0.4823 & \eu{0.7915} & \eb{0.7597} & \eb{0.7483} & \eb{0.7464} & 0.6193 \\
    \bottomrule
  \end{tabular}
  }
  \vspace{-0.2cm}
  \caption{\textbf{Evaluation of human-object interaction generation on the OMOMO~\cite{li2023object} dataset.} We train MaskedMimic~\cite{tessler2024maskedmimic} with different human motion datasets and then evaluate the object interaction ability by interpolating object motion and controlling human hands to follow the objects.
  }
  \vspace{-0.1cm}
  \label{tab:zero_shot_omomo}
\end{table*}
\begin{table*}
  \centering
  \resizebox{0.8\textwidth}{!}{
  \begin{tabular}{l|l|cc|ccccc}
    \toprule
    \multirow{2}{*}{Model} & \multirow{2}{*}{Training datasets} & \multicolumn{2}{c|}{Human motion} & \multicolumn{5}{c}{Interaction} \\ 
   & & $\text{H}_{\text{feet}}$ $\downarrow$ & FS $\downarrow$ & $\text{C}_{\text{prec}}$ $\uparrow$ & $\text{C}_{\text{rec}}$ $\uparrow$ & $\text{C}_{\text{f1}}$ $\uparrow$ & $\text{C}_{\text{percent}}$  & $\text{P}_{\text{hand}}$ $\downarrow$ \\
    \midrule
    CHOIS~\cite{li2023controllable} & OMOMO~\cite{li2023object} & 1.91 & 0.4546 & 0.6994 & 0.5813 & 0.5873 & 0.6296 & \eb{0.6889} \\
     \midrule
    \multirow{4}{*}{MaskedMimic~\cite{tessler2024maskedmimic}} & AMASS~\cite{mahmood2019amass} & -0.32 & \eb{0.3320} & \eb{0.7020} & 0.5737 & 0.5685 & 0.6454 & 0.8097 \\
     & OMOMO~\cite{li2023object} & \eb{-0.01} & 0.4468 & 0.6975 & 0.5510 & 0.5575 & 0.6195 & \eu{0.8025} \\
     & \ours & \eu{0.03} & \eu{0.4460} & 0.6977 & \eu{0.6191} & \eu{0.5988} & \eu{0.6992} & 0.8098 \\
     & \ours + AMASS~\cite{mahmood2019amass} & -0.31 & 0.4904 & \eu{0.7015} & \eb{0.6936} & \eb{0.6490} & \eb{0.7706} & 0.8887 \\
    \bottomrule
  \end{tabular}
  }
  \vspace{-0.2cm}
  \caption{\textbf{Evaluation of human-object interaction generation on the BEHAVE~\cite{bhatnagar22behave} dataset.} Similar to Table~\ref{tab:zero_shot_omomo}, we report the performance of the models in Table~\ref{tab:zero_shot_omomo} on the BEHAVE~\cite{bhatnagar22behave} dataset.}
  \vspace{-0.1cm}
  \label{tab:zero_shot_behave}
\end{table*}

\subsection{Application: Zero-shot HOI generation}
\label{subsec:zero_shot}

\smallskip
\noindent\textbf{Implementation details.}
To demonstrate the zero-shot HOI generation capability of \ours, we conduct experiments on the standard benchmark~\cite{li2023controllable}, where the model must generate human–object motion conditioned on sparse object waypoints.
For this task, we adopt the spatially controllable network MaskedMimic~\cite{tessler2024maskedmimic}. However, since MaskedMimic~\cite{tessler2024maskedmimic} is primarily trained on human-only datasets~\cite{mahmood2019amass} with limited interactive motion, its ability to follow object control points is constrained. To overcome this limitation, we retrain the model on \ours and evaluate its performance on HOI generation tasks.
We further assess human motion quality and interaction fidelity by comparing against the supervised HOI framework CHOIS~\cite{li2023controllable} on OMOMO~\cite{li2023object} and BEHAVE~\cite{bhatnagar22behave}, using the same object waypoints for fair comparison. Object trajectories and rotations are interpolated using PCHIP and Slerp, respectively, and contact points are sampled based on object mesh points to guide hand control during motion generation. 
To identify hand-object contact points, we deploy a simple heuristic and sample points on the object surface that have smaller distance to human hands.

\noindent\textbf{Metrics.}
For HOI generation, we follow CHOIS~\cite{li2023controllable} and measure foot height ($\text{H}_{\text{feet}}$) and foot sliding (FS) to obtain the human motion quality. Contact precision ($\text{C}_{\text{prec}}$), recall ($\text{C}_{\text{rec}}$), F1 score ($\text{C}_{\text{f1}}$), contact percentage ($\text{C}_{\text{percent}}$) and penetration score ($\text{P}_{\text{hand}}$) are also calculated to evaluate the interaction performance, following prior work~\cite{li2023controllable}. More details about the evaluation metrics are provided in Section~\ref{sec:supp_metrics} of the Appendix.

\noindent\textbf{Results.}
We compare our models with CHOIS~\cite{li2023controllable} on the HOI benchmarks presented in Tables~\ref{tab:zero_shot_omomo}-\ref{tab:zero_shot_behave}. MaskedMimic~\cite{tessler2024maskedmimic} trained on OMOMO~\cite{li2023object} obtains the worst results, because OMOMO~\cite{li2023object} is small-scale, and thus the model fails to generalize due to the gap between the GT and the interpolated hands' trajectory. While the model trained on the AMASS~\cite{mahmood2019amass} dataset can achieve promising human motion quality, but due to lack of interactive training behaviors,
the model fails to accurately follow the trajectories of interactive hands, resulting in suboptimal interaction performance. Incorporating \ours data substantially enhances interaction quality, particularly in terms of object contact metrics. Notably, even though CHOIS~\cite{li2023controllable} is trained in a fully-supervised manner on the OMOMO~\cite{li2023object} dataset, our models trained on \ours data still outperform CHOIS~\cite{li2023controllable} in contact quality under zero-shot generation. Furthermore, in evaluations on the BEHAVE~\cite{bhatnagar22behave} dataset, where all models are assessed in zero-shot settings, our models consistently exceed CHOIS~\cite{li2023controllable} across most metrics. The consistent performance across both OMOMO~\cite{li2023object} and BEHAVE~\cite{bhatnagar22behave} demonstrates the robust zero-shot generalization capability of models trained on \ours data.

\smallskip
\noindent\textbf{User study.} We conduct a user study to further evaluate and compare the zero-shot HOI performance of our method with the supervised HOI approach CHOIS~\cite{li2023controllable}. Specifically, we generate 20 HOI sequences using CHOIS~\cite{li2023controllable} and MaskedMimic~\cite{tessler2024maskedmimic} trained on a combination of \ours and AMASS~\cite{mahmood2019amass} (here denoted as Ours) for both the OMOMO~\cite{li2023object} and BEHAVE~\cite{bhatnagar22behave} datasets. Twenty participants were recruited to indicate their preference based on two criteria: (1)~the quality of human and object motion, and (2)~the accuracy of contact performance. As reported in Table~\ref{tab:user1}, the consistent results across participants confirm the effectiveness and robustness of our \ours dataset in facilitating high-quality interaction generation. 
Notably, the advantage becomes more evident on the out-of-distribution BEHAVE~\cite{bhatnagar22behave} dataset, which indicates the model trained on \ours could exhibit superior generalization and robustness in zero-shot interaction scenarios across distinct HOI datasets.

\begin{table} 
\centering\setlength{\tabcolsep}{8pt}
\resizebox{0.47\textwidth}{!}{
    \begin{tabular}{l|c|c}
        \toprule
        Methods & Preference on OMOMO~\cite{li2023object} (\%) & Preference on BEHAVE~\cite{bhatnagar22behave} (\%)  \\ 
       \midrule
        CHOIS~\cite{li2023controllable} & 38.5 & 23.5 \\ 
        Ours & \eb{61.5} & \eb{76.5} \\
       \bottomrule
    \end{tabular}
    }
    \vspace{-0.2cm}
    \caption{\textbf{User studies on OMOMO~\cite{li2023object} and BEHAVE~\cite{bhatnagar22behave} datasets.} "Ours" indicates trained on the combination of our \ours and AMASS~\cite{mahmood2019amass} datasets.
    }
    \vspace{-0.3cm}
    \label{tab:user1}
\end{table}

\begin{figure*}
  \centering
  \begin{subfigure}[b]{0.49\textwidth}
    \includegraphics[width=\linewidth]{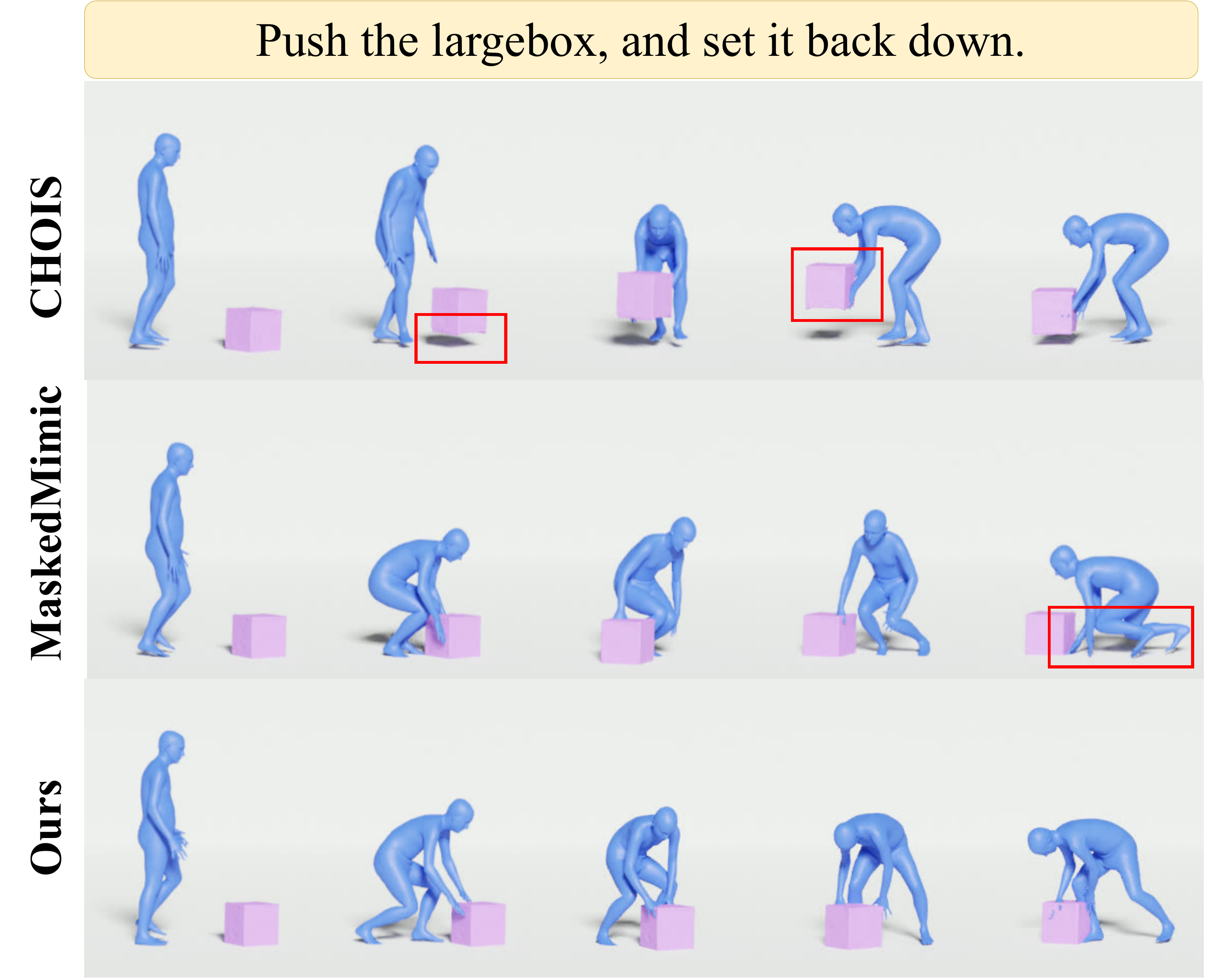}
    \caption{{Visual comparison on OMOMO~\cite{li2023object} dataset.}}
    \label{fig:compare_omomo}
  \end{subfigure}
  \hfill 
  \begin{subfigure}[b]{0.49\textwidth}
    \includegraphics[width=\linewidth]{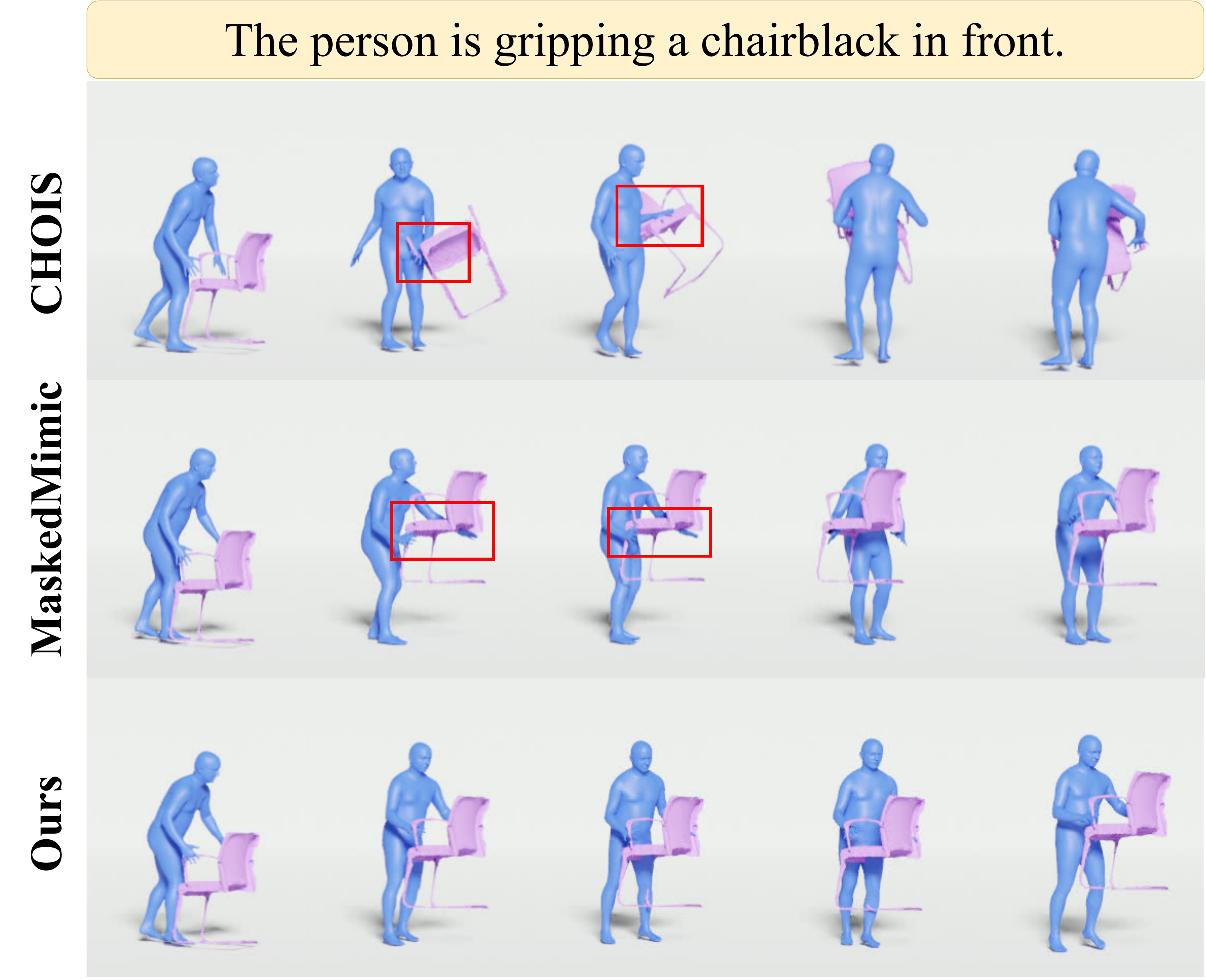}
    \caption{{Visual comparison on BEHAVE~\cite{bhatnagar22behave} datasets.}}
    \label{fig:compare_behave}
  \end{subfigure}
  \vspace{-0.2cm}
   \caption{\textbf{Visual comparison of CHOIS~\cite{li2023controllable}, MaskedMimic~\cite{tessler2024maskedmimic} and Ours when evaluated on OMOMO~\cite{li2023object} and BEHAVE~\cite{bhatnagar22behave} datasets.} Note that original MaskedMimic~\cite{tessler2024maskedmimic} is trained on AMASS~\cite{mahmood2019amass} dataset and Ours denotes MaskedMimic trained on the combination of \ours and AMASS~\cite{mahmood2019amass}. Red boxes highlight examples of unrealistic human action, unrealistic object motion or interaction.}
  \vspace{-0.1cm}
   \label{fig:compare_with_chois}
\end{figure*}

\smallskip
\noindent\textbf{Qualitative results.} We qualitatively compare CHOIS~\cite{li2023controllable} with zero-shot HOI generation using MaskedMimic~\cite{tessler2024maskedmimic}, and MaskedMimic trained on the combination of \ours and AMASS~\cite{mahmood2019amass}~(Ours). 
Evaluation is conducted on the HOI datasets OMOMO~\cite{li2023object} and BEHAVE~\cite{bhatnagar22behave}, as illustrated in Figure~\ref{fig:compare_with_chois}. From the example in Figure~\ref{fig:compare_omomo}, we observe that CHOIS~\cite{li2023controllable} generates unrealistic object motion (the object appears to float even when the person is distant) and the motion is not consistent with the textual description (e.g., the person is lifting rather than pushing the object). For the original MaskedMimic~\cite{tessler2024maskedmimic}, the person struggles to follow the object trajectory and nearly falls due to distributional shifts. In Figure~\ref{fig:compare_behave}, both CHOIS~\cite{li2023controllable} and MaskedMimic~\cite{tessler2024maskedmimic} fail to produce plausible contacts: CHOIS~\cite{li2023controllable} exhibits varying contact positions, whereas MaskedMimic~\cite{tessler2024maskedmimic} suffers from penetration issues. These examples demonstrate that a human motion generator trained on \ours results in interaction motions of higher quality and fidelity compared to other methods.
More visual results are provided in Figure~\ref{fig:more_vis_compare} of the Appendix.

\subsection{Application: Interaction in 3D Scenes}
\label{subsec:agent}

Previous section addresses the task of HOI generation given object waypoints and no scene information. 
Here we extend this approach to fully-automatic HOI generation and address multiple interaction types including collision-free navigation, multi-person collaboration as well as single and multi-object manipulation. Some methods, such as CHOIS~\cite{li2023controllable} and CooHOI~\cite{gao2024coohoi}, primarily focus on specific interaction types, whereas others, e.g., TokenHSI~\cite{pan2025tokenhsi}, support diverse interaction generation, but require manual task planning and transition handling, which is labor-intensive. 
To overcome these limitations, we formulate the problem as interaction generation within a given 3D environment guided by textual instructions. We introduce a LLM-based agent framework, {\bfseries \agent}, which performs automatic task planning and enables zero-shot generation of diverse interaction behaviors in complex 3D scenes.

\smallskip
\noindent\textbf{\agent framework.}
The \agent framework consists of a LLM-based high-level planner and a low-level motion generator. Given a human-level instruction and an environment state, including the position, orientation, and size of each object in the scene, the LLM planner analyzes spatial relationships and generates detailed step-wise plans along with collision-free human or object waypoints. As described in Section~\ref{subsec:zero_shot}, the sparse waypoints are then interpolated into full object motion. Based on the object motion, the contact points are sampled from the object surface to guide a human motion generator.
The low-level generator, MaskedMimic~\cite{tessler2024maskedmimic} trained on \ours, supports zero-shot HOI generation and can be conditioned on the generated waypoints to execute the desired interactions. 
In this manner, \agent enables flexible and robust interaction generation across complex 3D scenes. 
More detailed description, prompts and an illustrative example are provided in Section~\ref{sec:supp_agent} of the Appendix.

\begin{figure*}
  \centering
   \setlength{\tabcolsep}{8pt}
   \includegraphics[width=0.98\textwidth]{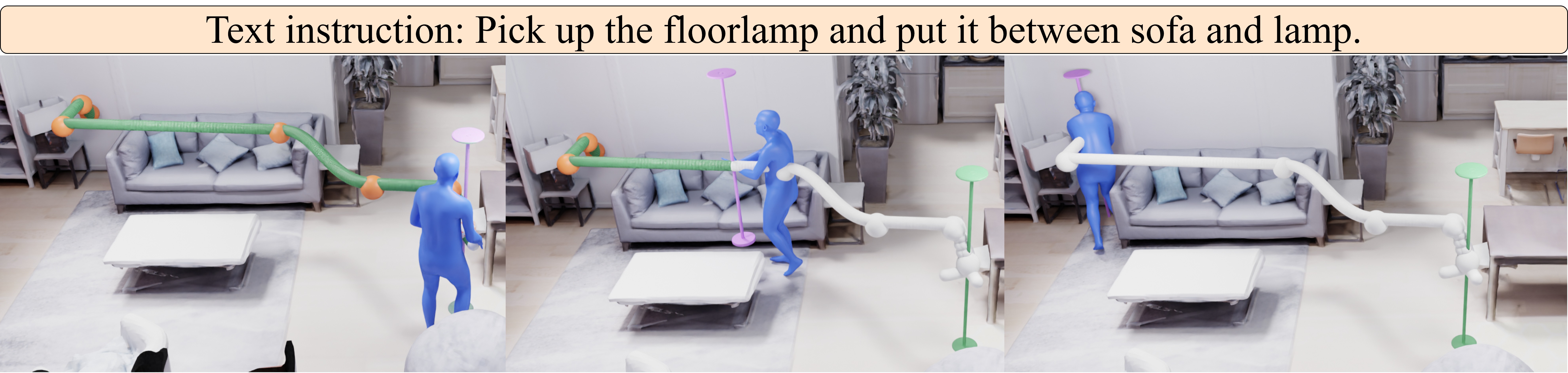}
   \vspace{-0.2cm}
   \caption{\textbf{Illustration of zero-shot human and object motion generation by our \agent.} Given an initial scene and text instruction, the LLM planner generates a detailed plan and collision-free sparse waypoints (orange). The contact points sampled from the object surface (green) are then used as hand control points for human motion generation.}
   \vspace{-0.3cm}
   \label{fig:agent_vis}
\end{figure*}

\smallskip
\noindent\textbf{Quantitative results.}
We evaluate our proposed \agent and use it to generate 20 HOI sequences in a 3D scene. The success rate is computed based on two criteria: (1)~the manipulated object reaches a position close to the target location, and (2)~the human remains upright without falling. Among the 20 generated sequences, 13 are successful, while the remaining failures are attributed to either wrong planning (3 cases) or the human subject losing balance and falling (4 cases).

\smallskip
\noindent\textbf{Qualitative results.} As illustrated in Figure~\ref{fig:agent_vis}, the high-level planner is able to generate instruction-following and scene-consistent waypoints used to control human motion generation. We also show the capabilities of our \agent in Figure~\ref{fig:teaser}. With our proposed \agent framework, we enable the automatic motion generation for various tasks:  collision-free human navigation, HOI in new 3D scenes, multi-object interaction and multi-person collaboration. 
More results can be found on the project page~\cite{interposewebpage}.

\section{Conclusion}
In this paper we proposed \ours, a large-scale and automatically created dataset with 3D human motions containing diverse human-object interactions. \ours enables both kinematics-based and physics-based models to generate higher quality motion sequences.
We further explored zero-shot interaction generation, and demonstrated advantages of our dataset compared to previous training setups. Moreover, we proposed \agent enabling fully-automatic zero-shot HOI generation in previously unseen complex 3D scenes. 

\smallskip
\noindent\textbf{Limitations and future work.}
First, \ours is currently composed of human motions only and excludes associated object movements.
With the progress in 3D object reconstruction, future work can augment \ours 
with the shape and motion of manipulated objects to provide additional data for HOI generation.  
Second, while our \agent enables high-level planning, collision-free trajectory generation, and interaction motion synthesis, its effectiveness depends on the underlying LLM and prompt engineering. Learning \agent could be another interesting direction for future work.
Overall, we believe that \ours and the proposed \agent framework will advance and inspire further research in HOI generation.

\smallskip\noindent\textbf{Acknowledgement.} This work was supported by the Institute of Information \& Communications Technology Planning \& Evaluation (IITP) grant funded by the Korean Government (MSIT) (No. RS-2024-00457882, National AI Research Lab Project).

{
    \small
    \bibliographystyle{ieeenat_fullname}
    \bibliography{references}

@String(PAMI = {IEEE Trans. Pattern Anal. Mach. Intell.})

@String(CVPR= {IEEE Conf. Comput. Vis. Pattern Recog.})

@String(ICCV= {Int. Conf. Comput. Vis.})

@String(ECCV= {Eur. Conf. Comput. Vis.})

@String(NIPS= {Adv. Neural Inform. Process. Syst.})

@String(TOG= {ACM Trans. Graph.})

@String(ACMMM= {ACM Int. Conf. Multimedia})

@String(ICLR = {Int. Conf. Learn. Represent.})

@String(AAAI = {AAAI})

@String(CVPRW= {IEEE Conf. Comput. Vis. Pattern Recog. Worksh.})

@String(PAMI  = {IEEE TPAMI})

@String(CVPR  = {CVPR})

@String(ICCV  = {ICCV})

@String(ECCV  = {ECCV})

@String(NIPS  = {NeurIPS})

@String(TOG   = {ACM TOG})

@String(ACMMM = {ACM MM})

@String(ICLR  = {ICLR})

@String(CVPRW= {CVPRW})

@string(NIPS={NeurIPS})

@string(AAAI={AAAI})

@string(TDV={3DV})

@misc{interposewebpage,
  title = {{InterPose} project webpage},
  howpublished = {\url{https://mael-zys.github.io/InterPose/}},
}

@inproceedings{liu2025hoigen,
  title={{HOIGen-1M}: A large-scale dataset for human-object interaction video generation},
  author={Liu, Kun and Liu, Qi and Liu, Xinchen and Li, Jie and Zhang, Yongdong and Luo, Jiebo and He, Xiaodong and Liu, Wu},
  booktitle=CVPR,
  year={2025}
}

@inproceedings{gao2024coohoi,
      title={{CooHOI}: Learning Cooperative Human-Object Interaction with Manipulated Object Dynamics},
      author={Gao, Jiawei and Wang, Ziqin and Xiao, Zeqi and Wang, Jingbo and Wang, Tai and Cao, Jinkun and Hu, Xiaolin and Liu, Si and Dai, Jifeng and Pang, Jiangmiao},
      booktitle=NIPS,
      year={2024}
}

@inproceedings{pan2025tokenhsi,
      title={{TokenHSI}: Unified synthesis of physical human-scene interactions through task tokenization},
      author={Pan, Liang and Yang, Zeshi and Dou, Zhiyang and Wang, Wenjia and Huang, Buzhen and Dai, Bo and Komura, Taku and Wang, Jingbo},
      booktitle=CVPR,
      year={2025}
}

@inproceedings{wu2022saga,
  title={{SAGA}: Stochastic whole-body grasping with contact},
  author={Wu, Yan and Wang, Jiahao and Zhang, Yan and Zhang, Siwei and Hilliges, Otmar and Yu, Fisher and Tang, Siyu},
  booktitle=ECCV,
  year={2022},
}

@inproceedings{taheri2022goal,
  title={{GOAL}: Generating {4D} whole-body motion for hand-object grasping},
  author={Taheri, Omid and Choutas, Vasileios and Black, Michael J and Tzionas, Dimitrios},
  booktitle=CVPR,
  year={2022}
}

@inproceedings{xu2025intermimic,
  title={{InterMimic}: Towards Universal Whole-Body Control for Physics-Based Human-Object Interactions},
  author={Xu, Sirui and Ling, Hung Yu and Wang, Yu-Xiong and Gui, Liangyan},
  booktitle=CVPR,
  year={2025}
}

@article{lugaresi2019mediapipe,
  title={{MediaPipe}: A Framework for Building Perception Pipelines},
  author={Lugaresi, Cameron and Tang, Jiuqiang and Nash, Hartwig and McClanahan, Chris and Zhu, Mark and Chang, Chuo-Ling and Yong, Ming Guang and Lee, Joe and Pang, Fan and Fu, Victor and others},
  year={2019},
  journal={arXiv},
}

@inproceedings{SINC:ICCV:2023,
  author={Athanasiou, Nikos and Petrovich, Mathis and Black, Michael J. and Varol, G{\"u}l},
  title={{SINC}: Spatial Composition of {3D} Human Motions for Simultaneous Action Generation},
  booktitle=ICCV,
  year = {2023}
}

@article{jang2022motion,
  title={{Motion Puzzle}: Arbitrary motion style transfer by body part},
  author={Jang, Deok-Kyeong and Park, Soomin and Lee, Sung-Hee},
  journal={ACM Transactions on Graphics (TOG)},
  year={2022},
}

@inproceedings{
  guo2024generative,
  title={Generative Human Motion Stylization in Latent Space},
  author={chuan guo and Yuxuan Mu and Xinxin Zuo and Peng Dai and Youliang Yan and Juwei Lu and Li Cheng},
  booktitle=ICLR,
  year={2024},
}

@inproceedings{zhang2023generating,
  title={Generating human motion from textual descriptions with discrete representations},
  author={Zhang, Jianrong and Zhang, Yangsong and Cun, Xiaodong and Zhang, Yong and Zhao, Hongwei and Lu, Hongtao and Shen, Xi and Shan, Ying},
  booktitle=CVPR,
  year={2023}
}

@misc{PySceneDetect,
  author       = {Brandon Castellano},
  title        = {PySceneDetect},
  year         = {2021},
  howpublished = {\url{https://github.com/Breakthrough/PySceneDetect}},
}

@article{Qwen2.5-VL,
  title={{Qwen2.5-VL} Technical Report},
  author={Bai, Shuai and Chen, Keqin and Liu, Xuejing and Wang, Jialin and Ge, Wenbin and Song, Sibo and Dang, Kai and Wang, Peng and Wang, Shijie and Tang, Jun and Zhong, Humen and Zhu, Yuanzhi and Yang, Mingkun and Li, Zhaohai and Wan, Jianqiang and Wang, Pengfei and Ding, Wei and Fu, Zheren and Xu, Yiheng and Ye, Jiabo and Zhang, Xi and Xie, Tianbao and Cheng, Zesen and Zhang, Hang and Yang, Zhibo and Xu, Haiyang and Lin, Junyang},
  journal={arXiv},
  year={2025}
}

@article{xu2023vitpose++,
  title={{ViTPose++}: Vision transformer for generic body pose estimation},
  author={Xu, Yufei and Zhang, Jing and Zhang, Qiming and Tao, Dacheng},
  journal=PAMI,
  year={2023},
}

@inproceedings{pavlakos2019expressive,
  title={Expressive body capture: {3D} hands, face, and body from a single image},
  author={Pavlakos, Georgios and Choutas, Vasileios and Ghorbani, Nima and Bolkart, Timo and Osman, Ahmed AA and Tzionas, Dimitrios and Black, Michael J},
  booktitle=CVPR,
  year={2019}
}

@inproceedings{shin2024wham,
  title={{WHAM}: Reconstructing world-grounded humans with accurate {3D} motion},
  author={Shin, Soyong and Kim, Juyong and Halilaj, Eni and Black, Michael J},
  booktitle=CVPR,
  year={2024}
}

@article{tessler2024maskedmimic,
    author = {Tessler, Chen and Guo, Yunrong and Nabati, Ofir and Chechik, Gal and Peng, Xue Bin},
    title = {{MaskedMimic}: Unified Physics-Based Character Control Through Masked Motion Inpainting},
    year = {2024},
    journal={ACM Transactions on Graphics (TOG)},
}

@inproceedings{pavlakos2024reconstructing,
  title={Reconstructing hands in {3D} with transformers},
  author={Pavlakos, Georgios and Shan, Dandan and Radosavovic, Ilija and Kanazawa, Angjoo and Fouhey, David and Malik, Jitendra},
  booktitle=CVPR,
  year={2024}
}

@inproceedings{sigurdsson2016hollywood,
  title={Hollywood in homes: Crowdsourcing data collection for activity understanding},
  author={Sigurdsson, Gunnar A and Varol, G{\"u}l and Wang, Xiaolong and Farhadi, Ali and Laptev, Ivan and Gupta, Abhinav},
  booktitle=ECCV,
  year={2016},
}

@inproceedings{xue2022advancing,
  title={Advancing high-resolution video-language representation with large-scale video transcriptions},
  author={Xue, Hongwei and Hang, Tiankai and Zeng, Yanhong and Sun, Yuchong and Liu, Bei and Yang, Huan and Fu, Jianlong and Guo, Baining},
  booktitle=CVPR,
  year={2022}
}

@article{carreira2019short,
  title={A short note on the kinetics-700 human action dataset},
  author={Carreira, Joao and Noland, Eric and Hillier, Chloe and Zisserman, Andrew},
  journal={arXiv},
  year={2019}
}

@inproceedings{lv2024himonewbenchmarkfullbody,
    title={{HIMO}: A New Benchmark for Full-Body Human Interacting with Multiple Objects}, 
    author={Xintao Lv and Liang Xu and Yichao Yan and Xin Jin and Congsheng Xu and Shuwen Wu and Yifan Liu and Lincheng Li and Mengxiao Bi and Wenjun Zeng and Xiaokang Yang},
    year={2024},
    booktitle=ECCV,
    
}

@inproceedings{dai2024motionlcm,
  title={{MotionLCM}: Real-time controllable motion generation via latent consistency model},
  author={Dai, Wenxun and Chen, Ling-Hao and Wang, Jingbo and Liu, Jinpeng and Dai, Bo and Tang, Yansong},
  booktitle=ECCV,
  year={2024},
}

@article{lin2023motion,
  title={{Motion-X}: A large-scale {3D} expressive whole-body human motion dataset},
  author={Lin, Jing and Zeng, Ailing and Lu, Shunlin and Cai, Yuanhao and Zhang, Ruimao and Wang, Haoqian and Zhang, Lei},
  journal=NIPS,
  year={2023}
}

@inproceedings{xu2025interact,
    title     = {{InterAct}: Advancing Large-Scale Versatile {3D} Human-Object Interaction Generation},
    author    = {Xu, Sirui and Li, Dongting and Zhang, Yucheng and Xu, Xiyan and Long, Qi and Wang, Ziyin and Lu, Yunzhi and Dong, Shuchang and Jiang, Hezi and Gupta, Akshat and Wang, Yu-Xiong and Gui, Liang-Yan},
    booktitle = CVPR,
    year      = {2025},
}

@article{lu2025humoto,
  title={{HUMOTO}: A {4D} Dataset of Mocap Human Object Interactions},
  author={Lu, Jiaxin and Huang, Chun-Hao Paul and Bhattacharya, Uttaran and Huang, Qixing and Zhou, Yi},
  journal={arXiv},
  year={2025}
}

@article{luo2024grasping,
  title={Grasping diverse objects with simulated humanoids},
  author={Luo, Zhengyi and Cao, Jinkun and Christen, Sammy and Winkler, Alexander and Kitani, Kris and Xu, Weipeng},
  journal={arXiv},
  year={2024}
}

@article{wu2024human,
  title={Human-object interaction from human-level instructions},
  author={Wu, Zhen and Li, Jiaman and Liu, C Karen},
  journal={arXiv},
  year={2024}
}

@article{xu2024interdreamer,
  title={{InterDreamer}: Zero-Shot Text to {3D} Dynamic Human-Object Interaction},
  author={Xu, Sirui and Wang, Ziyin and Wang, Yu-Xiong and Gui, Liang-Yan},
  journal={arXiv},
  year={2024}
}

@article{li2023controllable,
  title={Controllable human-object interaction synthesis},
  author={Li, Jiaman and Clegg, Alexander and Mottaghi, Roozbeh and Wu, Jiajun and Puig, Xavier and Liu, C Karen},
  journal={arXiv},
  year={2023}
}

@article{peng2023hoi,
  title={{HOI-Diff}: Text-driven synthesis of {3D} human-object interactions using diffusion models},
  author={Peng, Xiaogang and Xie, Yiming and Wu, Zizhao and Jampani, Varun and Sun, Deqing and Jiang, Huaizu},
  journal={arXiv},
  year={2023}
}

@inproceedings{diller2024cg,
  title={{CG-HOI}: Contact-guided {3D} human-object interaction generation},
  author={Diller, Christian and Dai, Angela},
  booktitle=CVPR,
  year={2024}
}

@inproceedings{xu2023interdiff,
  title={{InterDiff}: Generating {3D} human-object interactions with physics-informed diffusion},
  author={Xu, Sirui and Li, Zhengyuan and Wang, Yu-Xiong and Gui, Liang-Yan},
  booktitle=ICCV,
  year={2023}
}

@article{li2023object,
  title={Object motion guided human motion synthesis},
  author={Li, Jiaman and Wu, Jiajun and Liu, C Karen},
  journal={ACM Transactions on Graphics (TOG)},
  year={2023},
}

@inproceedings{bhatnagar22behave,
    title = {{BEHAVE}: Dataset and Method for Tracking Human Object Interactions},
    author={Bhatnagar, Bharat Lal and Xie, Xianghui and Petrov, Ilya and Sminchisescu, Cristian and Theobalt, Christian and Pons-Moll, Gerard},
    booktitle = CVPR,
    year = {2022},
    }

@inproceedings{braun2024physically,
  title={Physically plausible full-body hand-object interaction synthesis},
  author={Braun, Jona and Christen, Sammy and Kocabas, Muhammed and Aksan, Emre and Hilliges, Otmar},
  booktitle=TDV,
  year={2024}
}

@inproceedings{jiang2024scaling,
  title={Scaling up dynamic human-scene interaction modeling},
  author={Jiang, Nan and Zhang, Zhiyuan and Li, Hongjie and Ma, Xiaoxuan and Wang, Zan and Chen, Yixin and Liu, Tengyu and Zhu, Yixin and Huang, Siyuan},
  booktitle=CVPR,
  year={2024}
}

@inproceedings{jiang2024autonomous,
  title={Autonomous character-scene interaction synthesis from text instruction},
  author={Jiang, Nan and He, Zimo and Wang, Zi and Li, Hongjie and Chen, Yixin and Huang, Siyuan and Zhu, Yixin},
  booktitle={SIGGRAPH Asia},
  year={2024}
}

@inproceedings{kim2025parahome,
  title={{ParaHome}: Parameterizing everyday home activities towards {3D} generative modeling of human-object interactions},
  author={Kim, Jeonghwan and Kim, Jisoo and Na, Jeonghyeon and Joo, Hanbyul},
  booktitle=CVPR,
  year={2025}
}

@misc{lou2025zeroshothumanobjectinteractionsynthesis,
    title={Zero-Shot Human-Object Interaction Synthesis with Multimodal Priors}, 
    author={Yuke Lou and Yiming Wang and Zhen Wu and Rui Zhao and Wenjia Wang and Mingyi Shi and Taku Komura},
    year={2025},
    eprint={2503.20118},
    archivePrefix={arXiv},
    primaryClass={cs.GR},
    url={https://arxiv.org/abs/2503.20118}, 
}

@article{li2024zerohsi,
  title={{ZeroHSI}: Zero-Shot {4D} Human-Scene Interaction by Video Generation},
  author={Li, Hongjie and Yu, Hong-Xing and Li, Jiaman and Wu, Jiajun},
  journal={arXiv},
  year={2024}
}

@article{milacski2024ghost,
  title={{GHOST}: Grounded Human Motion Generation with Open Vocabulary Scene-and-Text Contexts},
  author={Milacski, Zolt{\'a}n {\'A} and Niinuma, Koichiro and Kawamura, Ryosuke and de la Torre, Fernando and Jeni, L{\'a}szl{\'o} A},
  journal={arXiv},
  year={2024}
}

@article{xie2023omnicontrol,
  title={{OmniControl}: Control any joint at any time for human motion generation},
  author={Xie, Yiming and Jampani, Varun and Zhong, Lei and Sun, Deqing and Jiang, Huaizu},
  journal={arXiv},
  year={2023}
}

@inproceedings{karunratanakul2023guided,
  title={Guided motion diffusion for controllable human motion synthesis},
  author={Karunratanakul, Korrawe and Preechakul, Konpat and Suwajanakorn, Supasorn and Tang, Siyu},
  booktitle=ICCV,
  year={2023}
}

@inproceedings{zhao2023synthesizing,
  title={Synthesizing diverse human motions in {3D} indoor scenes},
  author={Zhao, Kaifeng and Zhang, Yan and Wang, Shaofei and Beeler, Thabo and Tang, Siyu},
  booktitle=ICCV,
 
  year={2023}
}

@inproceedings{tseng2023edge,
  title={{EDGE}: Editable dance generation from music},
  author={Tseng, Jonathan and Castellon, Rodrigo and Liu, Karen},
  booktitle=CVPR,
  year={2023}
}

@inproceedings{zhu2023taming,
  title={Taming diffusion models for audio-driven co-speech gesture generation},
  author={Zhu, Lingting and Liu, Xian and Liu, Xuanyu and Qian, Rui and Liu, Ziwei and Yu, Lequan},
  booktitle=CVPR,
  year={2023}
}

@inproceedings{li2021audio2gestures,
  title={{Audio2Gestures}: Generating diverse gestures from speech audio with conditional variational autoencoders},
  author={Li, Jing and Kang, Di and Pei, Wenjie and Zhe, Xuefei and Zhang, Ying and He, Zhenyu and Bao, Linchao},
  booktitle=ICCV,
  year={2021}
}

@inproceedings{li2022danceformer,
  title={{DanceFormer}: Music conditioned {3D} dance generation with parametric motion transformer},
  author={Li, Buyu and Zhao, Yongchi and Zhelun, Shi and Sheng, Lu},
  booktitle=AAAI,
  year={2022}
}

@inproceedings{zhang2022couch,
  title={{COUCH}: Towards controllable human-chair interactions},
  author={Zhang, Xiaohan and Bhatnagar, Bharat Lal and Starke, Sebastian and Guzov, Vladimir and Pons-Moll, Gerard},
  booktitle=ECCV,
  year={2022},
}

@inproceedings{yuan2020dlow,
  title={{DLow}: Diversifying latent flows for diverse human motion prediction},
  author={Yuan, Ye and Kitani, Kris},
  booktitle=ECCV,
  year={2020},
}

@inproceedings{zhang2021we,
  title={We are more than our joints: Predicting how {3D} bodies move},
  author={Zhang, Yan and Black, Michael J and Tang, Siyu},
  booktitle=CVPR,
  year={2021}
}

@inproceedings{ghosh2023imos,
  title={{IMoS}: Intent-Driven Full-Body Motion Synthesis for Human-Object Interactions},
  author={Ghosh, Anindita and Dabral, Rishabh and Golyanik, Vladislav and Theobalt, Christian and Slusallek, Philipp},
  booktitle={Computer Graphics Forum},
  year={2023},

}

@article{wang2022humanise,
  title={{HUMANISE}: Language-conditioned human motion generation in {3D} scenes},
  author={Wang, Zan and Chen, Yixin and Liu, Tengyu and Zhu, Yixin and Liang, Wei and Huang, Siyuan},
  journal=NIPS,
  year={2022}
}

@inproceedings{petrovich21actor,
    title = {Action-Conditioned 3{D} Human Motion Synthesis with Transformer {VAE}},
    author = {Petrovich, Mathis and Black, Michael J. and Varol, G{\"u}l},
    booktitle = ICCV,
    year = {2021}
}

@inproceedings{petrovich22temos,
  title     = {{TEMOS}: Generating diverse human motions from textual descriptions},
  author    = {Petrovich, Mathis and Black, Michael J. and Varol, Gul},
  booktitle = ECCV,
  year      = {2022}
}

@inproceedings{guo2020action2motion,
  title={{Action2Motion}: Conditioned generation of {3D} human motions},
  author={Guo, Chuan and Zuo, Xinxin and Wang, Sen and Zou, Shihao and Sun, Qingyao and Deng, Annan and Gong, Minglun and Cheng, Li},
  booktitle=ACMMM,
  year={2020}
}

@article{loper2015smpl,
  title={{SMPL}: A skinned multi-person linear model},
  author={Loper, Matthew and Mahmood, Naureen and Romero, Javier and Pons-Moll, Gerard and Black, Michael J},
  journal={ACM transactions on graphics (TOG)},
  year={2015},
}

@inproceedings{guo2022generating,
  title={Generating Diverse and Natural {3D} Human Motions From Text},
  author={Guo, Chuan and Zou, Shihao and Zuo, Xinxin and Wang, Sen and Ji, Wei and Li, Xingyu and Cheng, Li},
  booktitle=CVPR,
  year={2022}
}

@article{tevet2022MDM,
  title={Human Motion Diffusion Model},
  author={Tevet, Guy and Raab, Sigal and Gordon, Brian and Shafir, Yonatan and Bermano, Amit H and Cohen-Or, Daniel},
  journal={arXiv},
  year={2022}
}

@inproceedings{TEACH:3DV:2022, 
  title = {{TEACH}: {T}emporal {A}ction {C}ompositions for {3D} {H}umans},
  author = {Athanasiou, Nikos and Petrovich, Mathis and Black, Michael J. and Varol, G{\"u}l},
  booktitle = {{International Conference on {3D} Vision (3DV)}},
  year = {2022} 
  }

@inproceedings{mahmood2019amass,
  title={{AMASS}: Archive of motion capture as surface shapes},
  author={Mahmood, Naureen and Ghorbani, Nima and Troje, Nikolaus F and Pons-Moll, Gerard and Black, Michael J},
  booktitle=ICCV,
  year={2019}
}

@inproceedings{taheri2020grab,
  title={{GRAB}: A dataset of whole-body human grasping of objects},
  author={Taheri, Omid and Ghorbani, Nima and Black, Michael J and Tzionas, Dimitrios},
  booktitle=ECCV,
  year={2020}
}

@InProceedings{guo2023momask,
  title={{MoMask}: Generative Masked Modeling of {3D} Human Motions},
  author={Guo, Chuan and Mu, Yuxuan and Javed, Muhammad Gohar and Wang, Sen and Cheng, Li},
  booktitle = CVPR,
  year={2024}
}

@article{petrovich2024multi,
  title={Multi-Track Timeline Control for Text-Driven {3D} Human Motion Generation},
  author={Petrovich, Mathis and Litany, Or and Iqbal, Umar and Black, Michael J and Varol, G{\"u}l and Peng, Xue Bin and Rempe, Davis},
  journal=CVPRW,
  year={2024}
}
}

\appendix

\renewcommand{\thefigure}{A.\arabic{figure}} 
\setcounter{figure}{0} 
\renewcommand{\thetable}{A.\arabic{table}}
\setcounter{table}{0} 

\clearpage
\setcounter{page}{1}
\maketitlesupplementaryarxiv

In this appendix, we provide:
\begin{itemize}
\item Section~\ref{sec:supp_metrics}: details on the evaluation metrics for the HOI task.
\item Section~\ref{sec:supp_exp}: results of additional experiments and analyses.
\item Section~\ref{sec:supp_dataset}: further examples of motion sequences and the full text annotation prompt.
\item Section~\ref{sec:supp_agent}: a detailed description of the \agent framework, including prompts and a sample execution log.
\item Additional details and qualitative results are available on the project page~\cite{interposewebpage}.
\end{itemize}

\section{Evaluation metrics for HOI generation task}
\label{sec:supp_metrics}

We further explain the interaction metrics used in prior works~\cite{li2023object,li2023controllable} for HOI task.

\smallskip
\noindent\textbf{Penetration score.} 
At each time step $t$, we evaluate the signed distance function (SDF) of the object for every vertex $i$ on the reconstructed human mesh to obtain the signed distance $d_i^t$. 
A collision event is detected if there exists any vertex satisfying $d_i^t < 0$ and $|d_i^t| > 4\,\mathrm{cm}$. 
By accumulating such events across the sequence, we compute the overall collision percentage. 

\smallskip
\noindent\textbf{Contact performance.} 
To assess contact accuracy, we employ precision ($C_\mathrm{prec}$), recall ($C_\mathrm{rec}$), and F1 score ($C_\mathrm{F1}$) from the object detection evaluation. 
For each frame, we calculate the Euclidean distance between hand positions and the object mesh, and assign contact labels using a $5\,\mathrm{cm}$ threshold. 
The same procedure is applied to the ground-truth hand positions, enabling the determination of true positives, false positives, and false negatives for computing $C_\mathrm{prec}$, $C_\mathrm{rec}$, and $C_\mathrm{F1}$. The final metrics are obtained by first computing these scores on a per-sequence basis and then averaging them over all sequences.

\section{Experimental analysis}
\label{sec:supp_exp}

\subsection{Additional spatial control experiment}

In addition to the benchmarks reported in Table~\ref{tab:spatial_control_mask}, we further evaluate our method on the multi-object interaction dataset HIMO~\cite{lv2024himonewbenchmarkfullbody}, as shown in Table~\ref{tab:spatial_control_mask_himo}. Since most HIMO data involve loco-manipulation and its primary challenge lies in fine-grained finger manipulation of small objects, the benchmark measures hand control accuracy at relatively loose thresholds (0.5 m and 0.2 m), which can lead to saturated results. Nevertheless, our model trained on the combination of AMASS~\cite{mahmood2019amass} and \ours consistently outperforms the alternatives.

\subsection{Detailed results of dataset size impact}
Figure~\ref{fig:dataset} in the main paper illustrates the effect of dataset size, while Table~\ref{tab:ablation_data_ratio} provides the corresponding quantitative results. We observe that incorporating more data from \ours significantly enhances spatial controllability compared to OmniControl~\cite{xie2023omnicontrol} trained solely on HumanML3D~\cite{guo2022generating}. Notably, the improvement scales consistently as the proportion of \ours increases, with particularly strong gains on the HOI benchmarks OMOMO~\cite{li2023object} and BEHAVE~\cite{bhatnagar22behave}. These results suggest that larger-scale interaction datasets can further drive progress in HOI research.

\begin{table}
  \centering
  \resizebox{0.47\textwidth}{!}{
  \begin{tabular}{l|c|ccc}
    \toprule
    \multirow{2}{*}{Control joints} & \multirow{2}{*}{Training datasets} & \multicolumn{3}{c}{HIMO~\cite{lv2024himonewbenchmarkfullbody}} \\ 
     & & Succ 0.5 $\uparrow$ & Succ 0.2 $\uparrow$ & MPJPE $\downarrow$ \\
    \midrule
    \multirow{4}{*}{Pelvis} & AMASS~\cite{mahmood2019amass} & \eu{0.9985} & 0.9879 & \eu{25.1} \\
    & OMOMO~\cite{li2023object} & 0.9645 & 0.7870 & 101.6 \\
     & \ours & 0.9920 & \eu{0.9896} & 35.9 \\
     & AMASS\cite{mahmood2019amass} + \ours & \eb{0.9997} & \eb{0.9988} & \eb{22.9} \\
     \midrule
    \multirow{4}{*}{Hands} & AMASS~\cite{mahmood2019amass} & \eu{0.9941} & 0.9627 & 60.8 \\
     & OMOMO~\cite{li2023object} & 0.8513 & 0.7337 & 262.8 \\
     & \ours & 0.9899 & \eu{0.9787} & \eu{58.4} \\
     & AMASS\cite{mahmood2019amass} + \ours & \eb{0.9997} & \eb{0.9947} & \eb{44.1} \\
    \bottomrule
  \end{tabular}
  }
  \caption{\textbf{Spatial controllability of human motion generation using MaskedMimic~\cite{tessler2024maskedmimic}.} We train MaskedMimic~\cite{tessler2024maskedmimic} with different datasets and then evaluate all the models with different control settings on object interaction datasets (HIMO~\cite{lv2024himonewbenchmarkfullbody}). The best and second-best results are indicated with bold and underlined font respectively.}
  \label{tab:spatial_control_mask_himo}
\end{table}

\begin{table*}
  \centering
  \resizebox{0.99\textwidth}{!}{
  \begin{tabular}{l|c|cccc|cccc|cccc}
    \toprule
    \multirow{2}{*}{Control joints} & \multirow{2}{*}{Training datasets} & \multicolumn{4}{c|}{HumanML3D~\cite{guo2022generating}} & \multicolumn{4}{c|}{OMOMO~\cite{li2023object}} & \multicolumn{4}{c}{BEHAVE\cite{bhatnagar22behave}} \\ 
     & & FS $\downarrow$ & Traj err 0.5 $\downarrow$ & Traj err 0.2 $\downarrow$ & MPJPE $\downarrow$ & FS $\downarrow$ & Traj err 0.5 $\downarrow$ & Traj err 0.2 $\downarrow$ & MPJPE $\downarrow$ & FS $\downarrow$ & Traj err 0.5 $\downarrow$ & Traj err 0.2 $\downarrow$ & MPJPE $\downarrow$ \\
    \midrule
    \multirow{5}{*}{Pelvis} & HumanML3D~\cite{guo2022generating} & \eu{0.0585} & 0.0957 & 0.3467 & 0.0708 & \eu{0.0945} & 0.0604 & 0.5208 & 0.0836 & 0.0497 & 0.0179 & 0.3259 & 0.0668 \\
     & HumanML3D~\cite{guo2022generating} + 10\% \ours & 0.0666 & 0.0703 & 0.2910 & 0.0610 & 0.1157 & 0.0354 & 0.3479 & 0.0608 & 0.0667 & 0.0223 & 0.2143 & 0.0576 \\
     & HumanML3D~\cite{guo2022generating} + 20\% \ours & 0.0629 & 0.0693 & 0.2656 & 0.0570 & 0.1073 & \eu{0.0271} & 0.3187 & 0.0561 & 0.0548 & \eu{0.0217} & 0.1339 & 0.0417 \\
     & HumanML3D~\cite{guo2022generating} + 50\% \ours & \eb{0.0480} & \eu{0.0635} & \eu{0.2305} & \eu{0.0508} & \eb{0.0922} & \eb{0.0042} & \eu{0.1604} & \eu{0.0409} & \eb{0.0476} & \eb{0} & \eu{0.0804} & \eu{0.0270} \\
     & HumanML3D~\cite{guo2022generating} + 100\% \ours & 0.0656 & \eb{0.0508} & \eb{0.2207} & \eb{0.0489} & 0.1185 & \eb{0.0042} & \eb{0.1167} & \eb{0.0327} & \eu{0.0488} & \eb{0} & \eb{0.0402} & \eb{0.0232} \\
     \midrule
    \multirow{5}{*}{Hands} & HumanML3D~\cite{guo2022generating} & 0.0681 & 0.1377 & 0.5645 & 0.0849 & 0.0960 & 0.2313 & 0.8646 & 0.1275 & 0.0878 & 0.0670 & 0.6741 & 0.0911 \\
     & HumanML3D~\cite{guo2022generating} + 10\% \ours & 0.0654 & 0.1104 & 0.4971 & 0.0757 & 0.0944 & 0.1812 & 0.7812 & 0.1097 & 0.0879 & 0.0491 & 0.5759 & 0.0680 \\
     & HumanML3D~\cite{guo2022generating} + 20\% \ours & 0.0670 & 0.0957 & 0.4404 & 0.0683 & 0.0914 & 0.1083 & 0.6333 & 0.0889 & 0.0704 & 0.0357 & 0.4286 & 0.0554 \\
     & HumanML3D~\cite{guo2022generating} + 50\% \ours & \eb{0.0442} & \eu{0.0771} & \eu{0.3789} & \eu{0.0596} & \eu{0.0723} & \eu{0.0563} & \eu{0.4354} & \eu{0.0595} & \eb{0.0560} & \eu{0.0134} & \eu{0.2589} & \eu{0.0381} \\
     & HumanML3D~\cite{guo2022generating} + 100\% \ours & \eu{0.0524} & \eb{0.0605} & \eb{0.3320} & \eb{0.0555} & \eb{0.0741} & \eb{0.0354} & \eb{0.3396} & \eb{0.0461} & \eu{0.0639} & \eb{0.0045} & \eb{0.2098} & \eb{0.0320} \\
    \bottomrule
  \end{tabular}
  }
  \caption{Ablation on different data ratio with generative method: OmniControl~\cite{xie2023omnicontrol}.}
  \label{tab:ablation_data_ratio}
\end{table*}

\subsection{Visual comparison of HOI generation}

\begin{figure*}
  \centering
   \begin{subfigure}[b]{0.98\textwidth}
    \includegraphics[width=\linewidth]{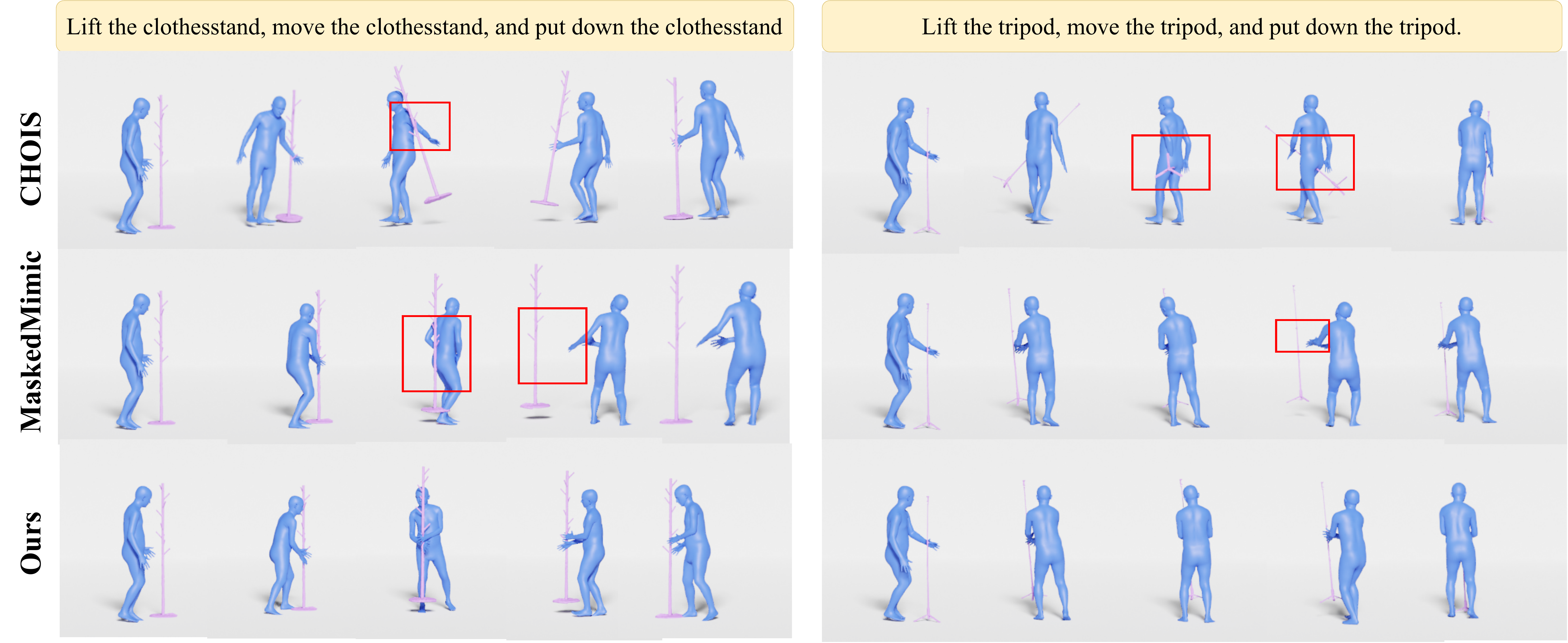}
    \caption{Visual comparison on OMOMO~\cite{li2023object} dataset.}
  \end{subfigure}
  \hfill 
  \begin{subfigure}[b]{0.98\textwidth}
    \includegraphics[width=\linewidth]{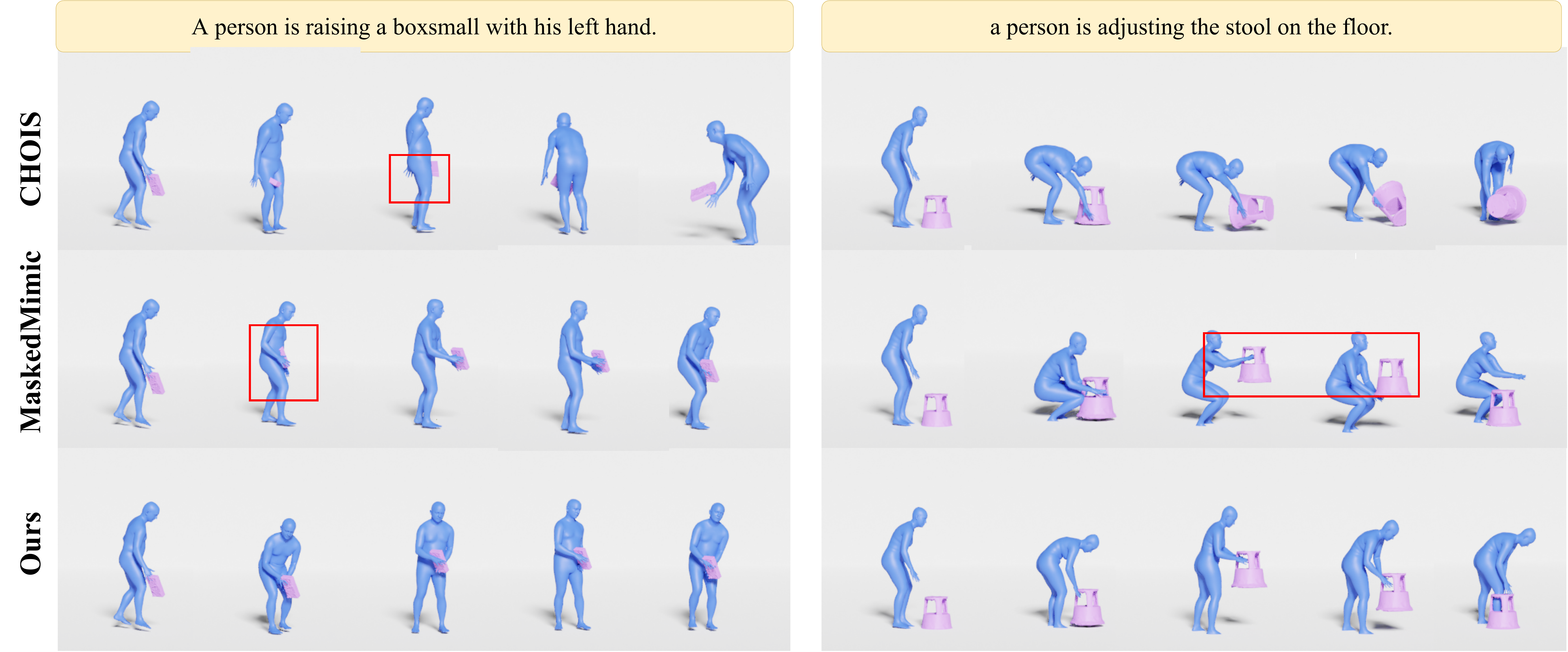}
    \caption{Visual comparison on BEHAVE~\cite{bhatnagar22behave} dataset}
  \end{subfigure}
   \caption{\textbf{Visual comparison of CHOIS~\cite{li2023controllable}, MaskedMimic~\cite{tessler2024maskedmimic} and MaskedMimic trained on \ours and AMASS (Ours) when evaluated on OMOMO~\cite{li2023object} and BEHAVE~\cite{bhatnagar22behave} datasets.} Similar to Figure~\ref{fig:compare_with_chois} in the main paper, we add more visual comparison on HOI generation.}
   \label{fig:more_vis_compare}
\end{figure*}

In addition to Figure~\ref{fig:compare_with_chois} in the main paper, Figure~\ref{fig:more_vis_compare} presents additional examples demonstrating both the performance improvement and the zero-shot capability enabled by the \ours dataset. We observe that the CHOIS~\cite{li2023controllable} model often produces unrealistic object motions that penetrate the human body, while MaskedMimic~\cite{tessler2024maskedmimic} trained on AMASS~\cite{mahmood2019amass} struggles to accurately follow interactive object trajectories, leading to penetration and loss of contact. In contrast, MaskedMimic trained on \ours exhibits stronger generalization to diverse hand–object interaction patterns, enabling the generation of motions with substantially improved contact quality.

\section{Details of \ours dataset}
\label{sec:supp_dataset}

\subsection{Motion examples}

\begin{figure*}
  \centering
   \begin{subfigure}[b]{0.49\textwidth}
    \includegraphics[width=\linewidth]{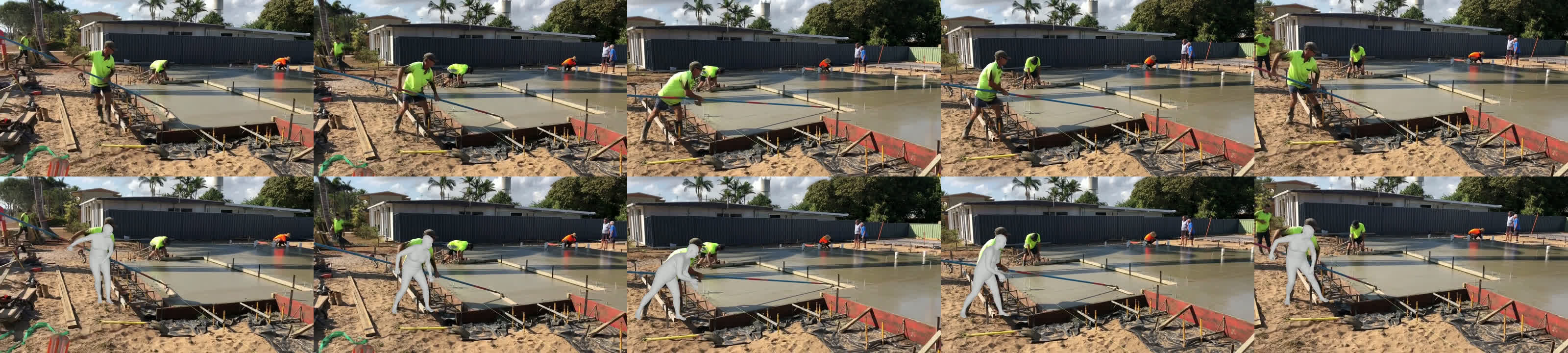}
  \end{subfigure}
  \hfill 
  \begin{subfigure}[b]{0.49\textwidth}
    \includegraphics[width=\linewidth]{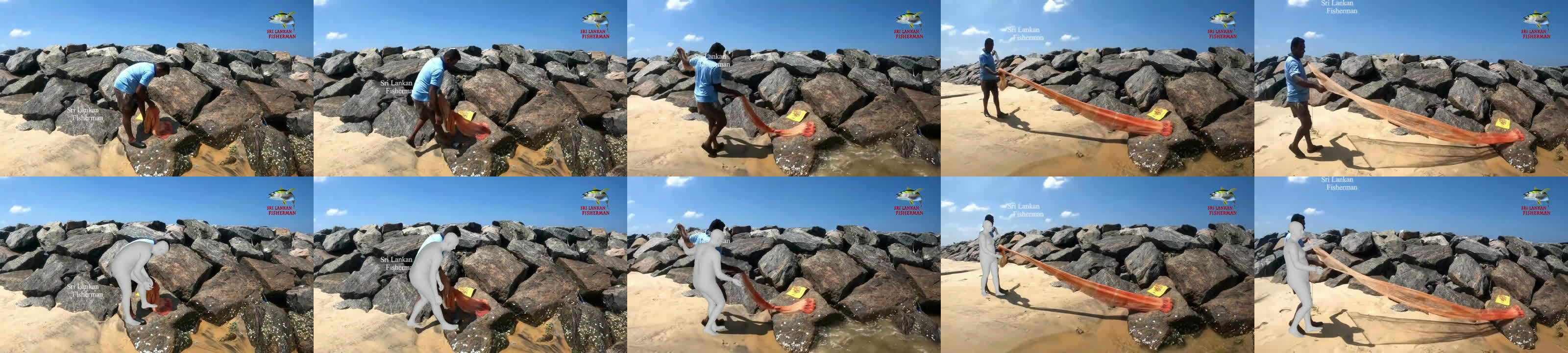}
  \end{subfigure}
  \hfill 
  \begin{subfigure}[b]{0.49\textwidth}
    \includegraphics[width=\linewidth]{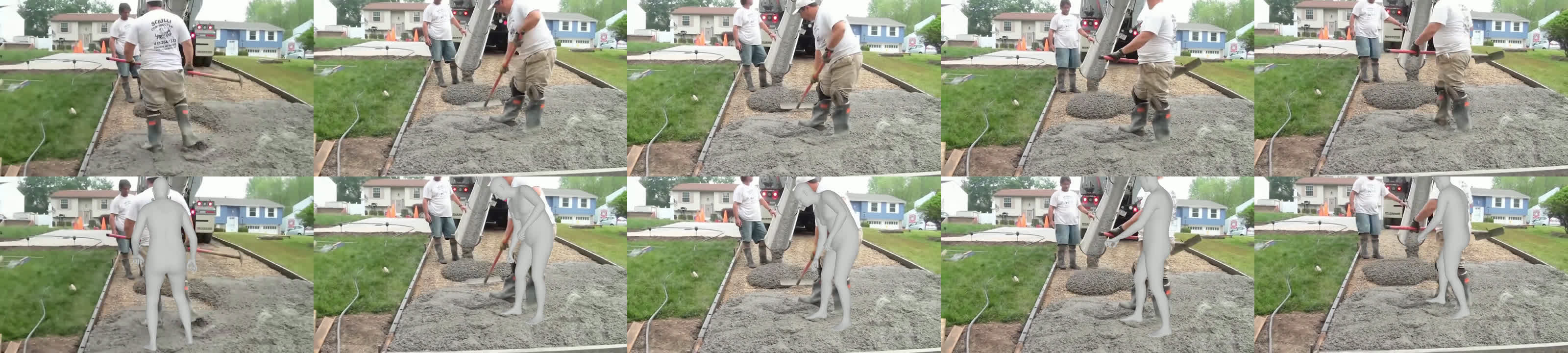}
  \end{subfigure}
   \hfill 
  \begin{subfigure}[b]{0.49\textwidth}
    \includegraphics[width=\linewidth]{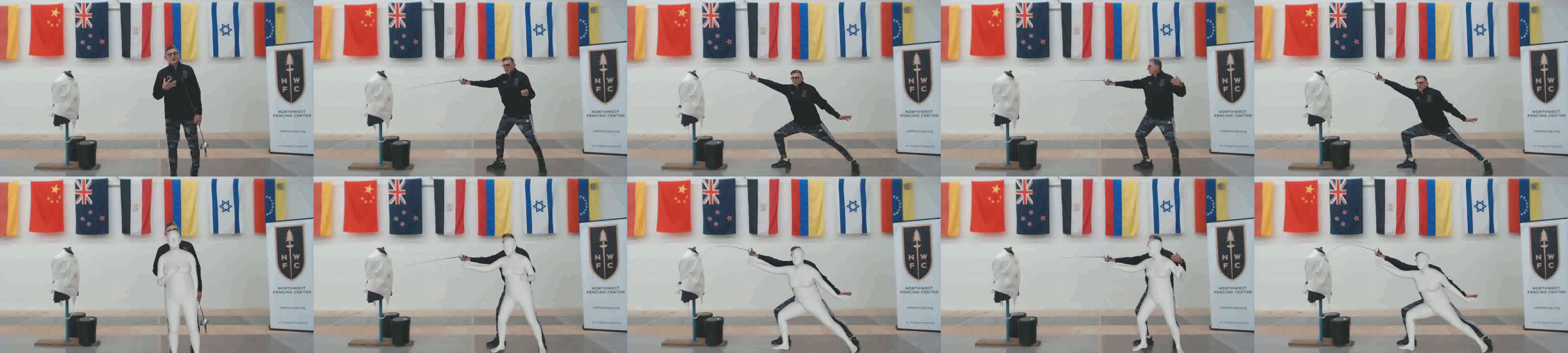}
  \end{subfigure}
   \hfill 
  \begin{subfigure}[b]{0.49\textwidth}
    \includegraphics[width=\linewidth]{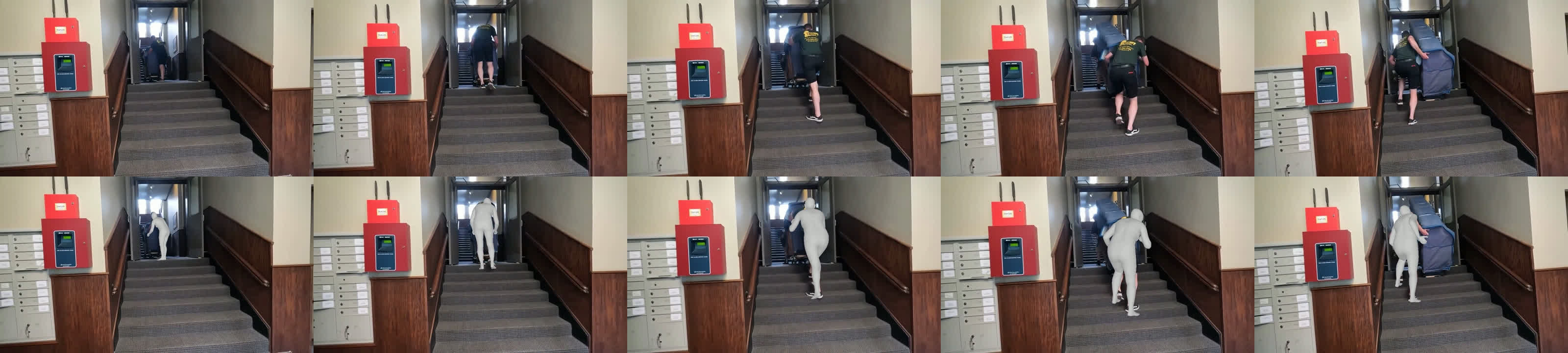}
  \end{subfigure}
  \hfill 
  \begin{subfigure}[b]{0.49\textwidth}
    \includegraphics[width=\linewidth]{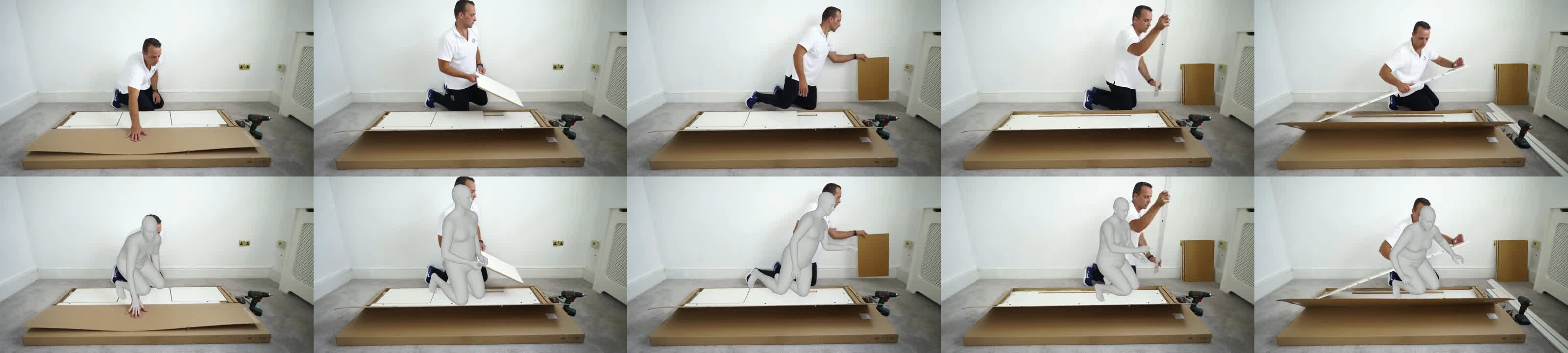}
  \end{subfigure}
  \hfill 
  \begin{subfigure}[b]{0.49\textwidth}
    \includegraphics[width=\linewidth]{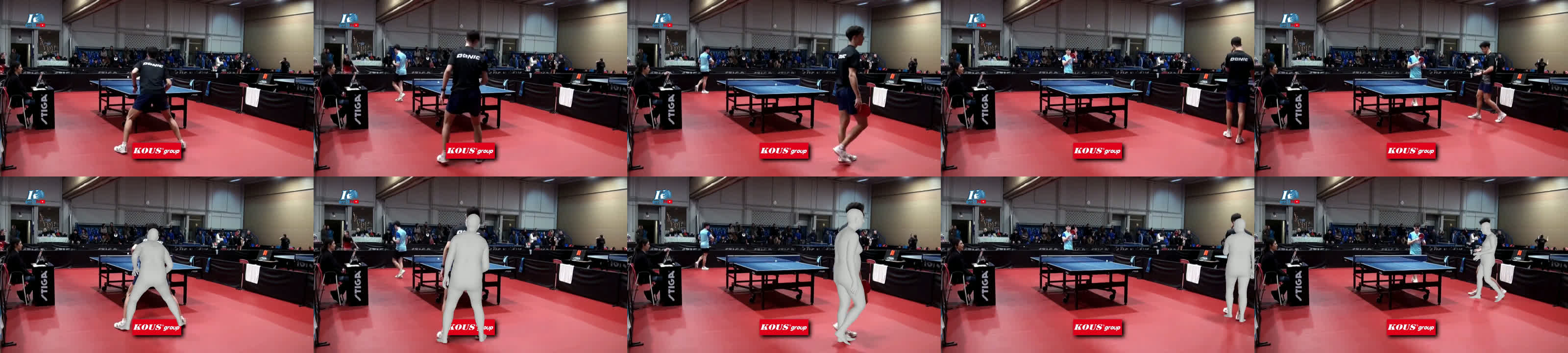}
  \end{subfigure}
  \hfill 
  \begin{subfigure}[b]{0.49\textwidth}
    \includegraphics[width=\linewidth]{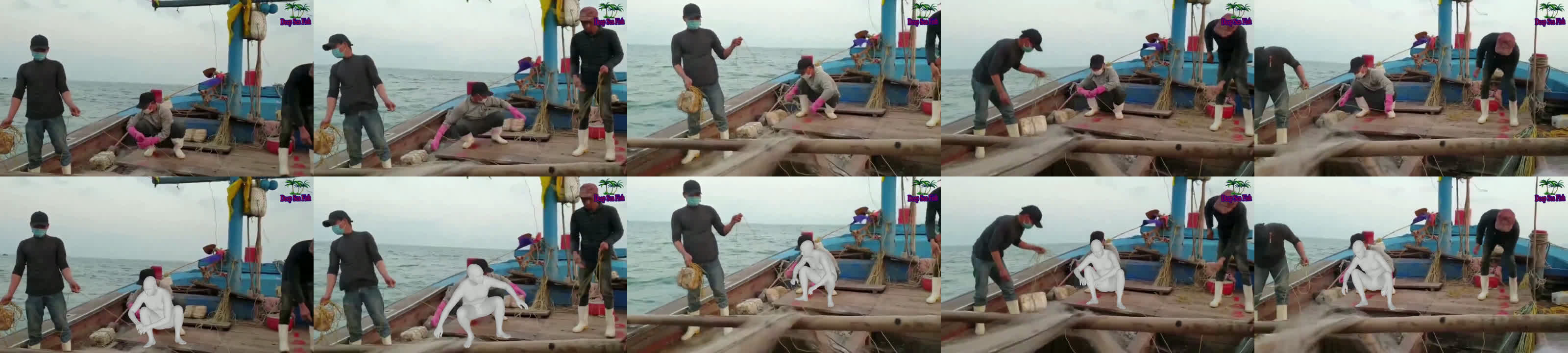}
  \end{subfigure}
  \hfill 
  \begin{subfigure}[b]{0.49\textwidth}
    \includegraphics[width=\linewidth]{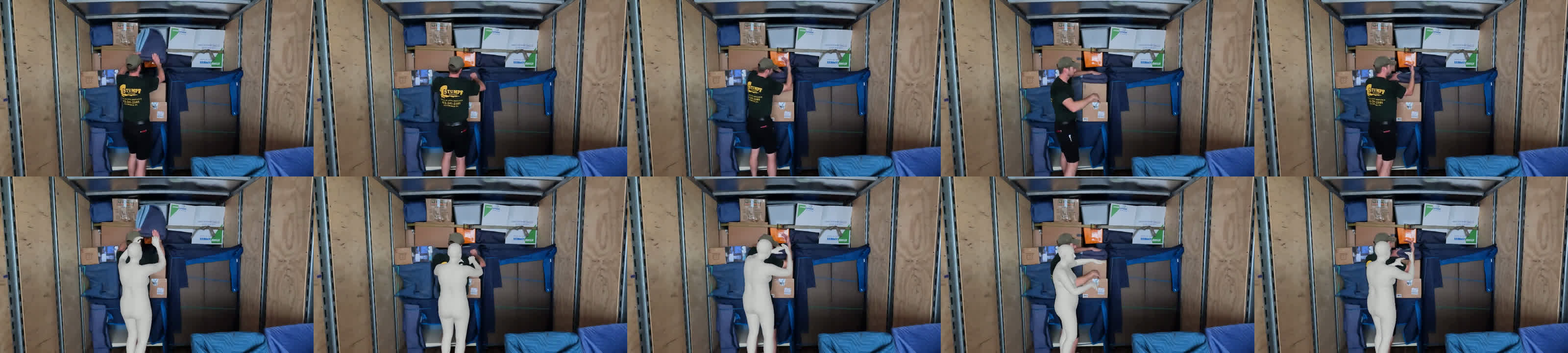}
  \end{subfigure}
  \hfill 
  \begin{subfigure}[b]{0.49\textwidth}
    \includegraphics[width=\linewidth]{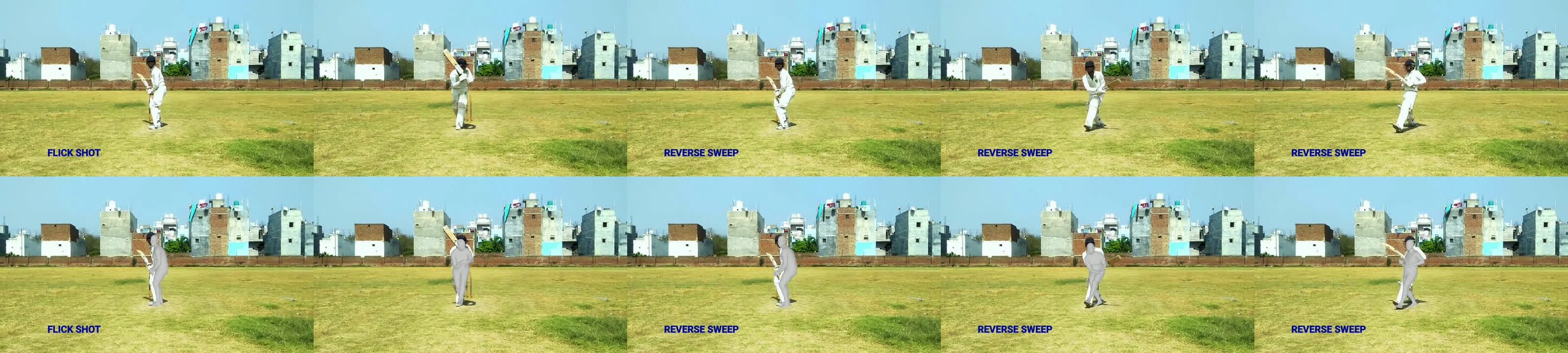}
  \end{subfigure}
  \hfill 
  \begin{subfigure}[b]{0.49\textwidth}
    \includegraphics[width=\linewidth]{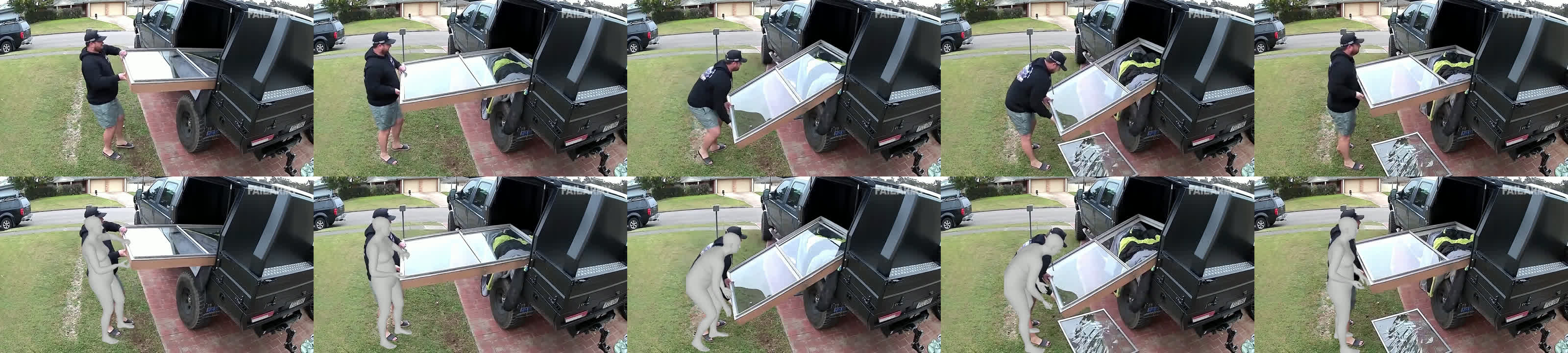}
  \end{subfigure}
  \hfill 
  \begin{subfigure}[b]{0.49\textwidth}
    \includegraphics[width=\linewidth]{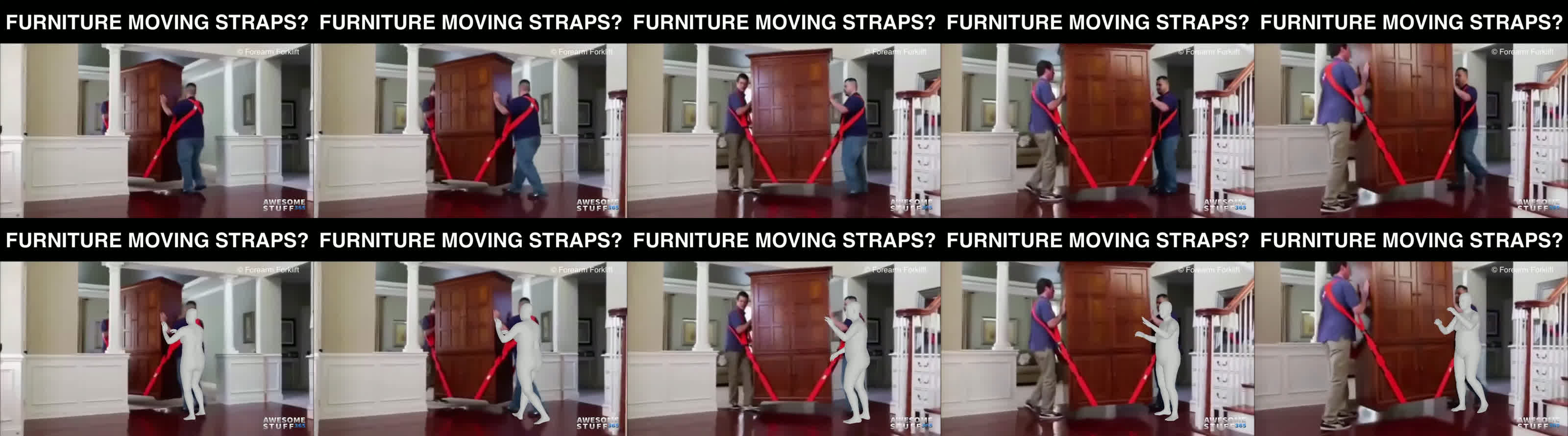}
  \end{subfigure}
  \hfill 
  \begin{subfigure}[b]{0.33\textwidth}
    \includegraphics[width=\linewidth]{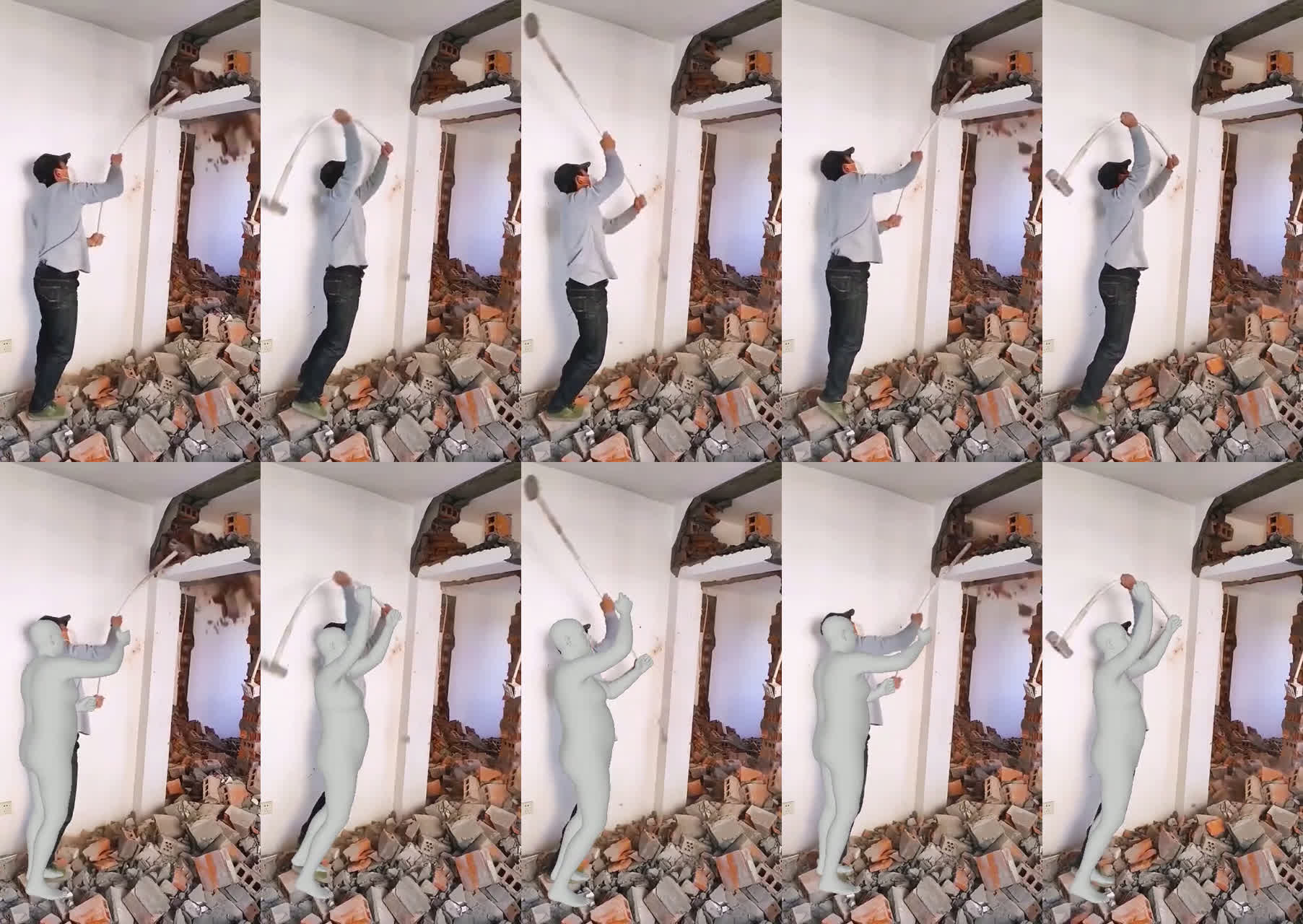}
  \end{subfigure}
  \hfill 
  \begin{subfigure}[b]{0.33\textwidth}
    \includegraphics[width=\linewidth]{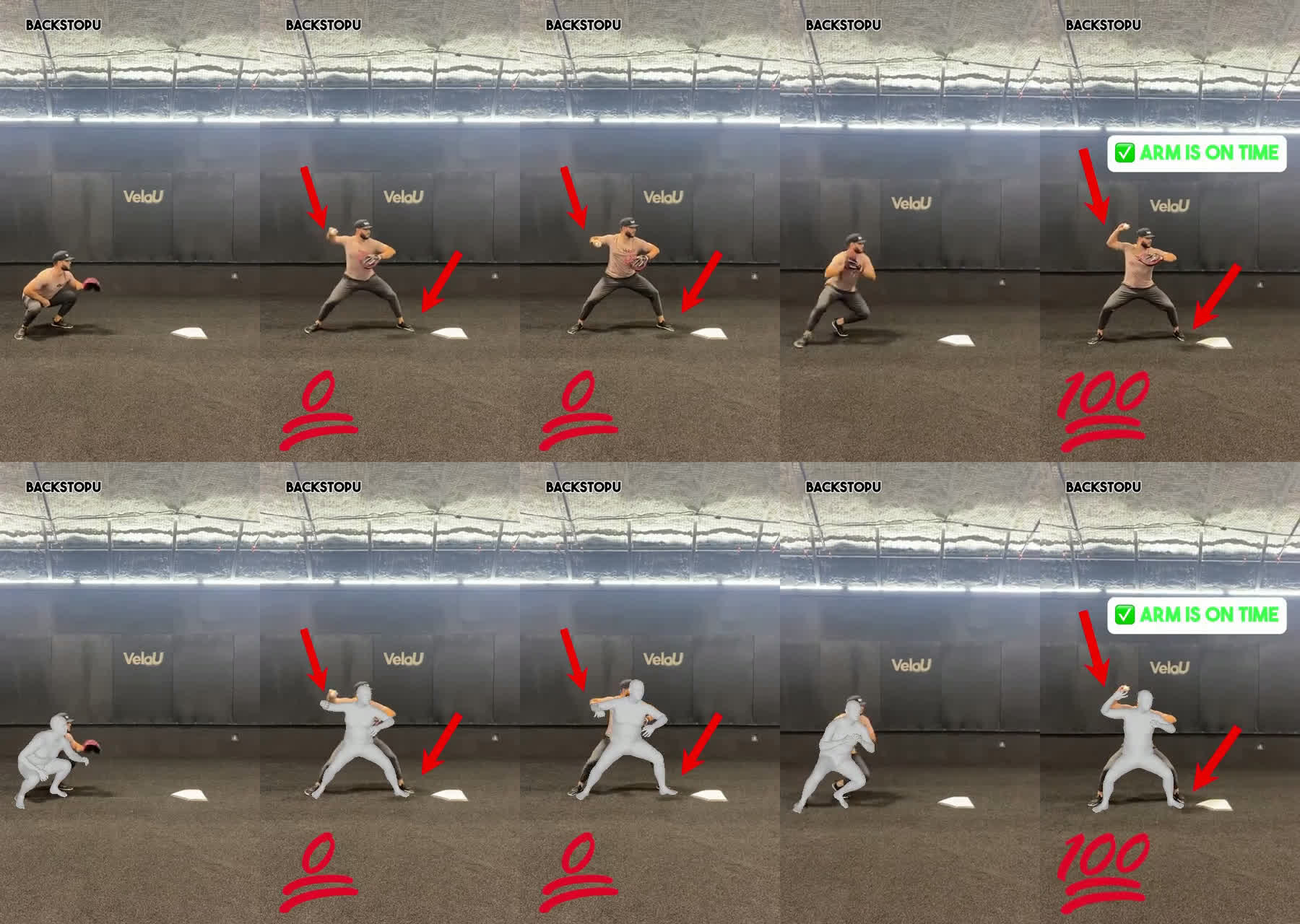}
  \end{subfigure}
  \hfill 
  \begin{subfigure}[b]{0.33\textwidth}
    \includegraphics[width=\linewidth]{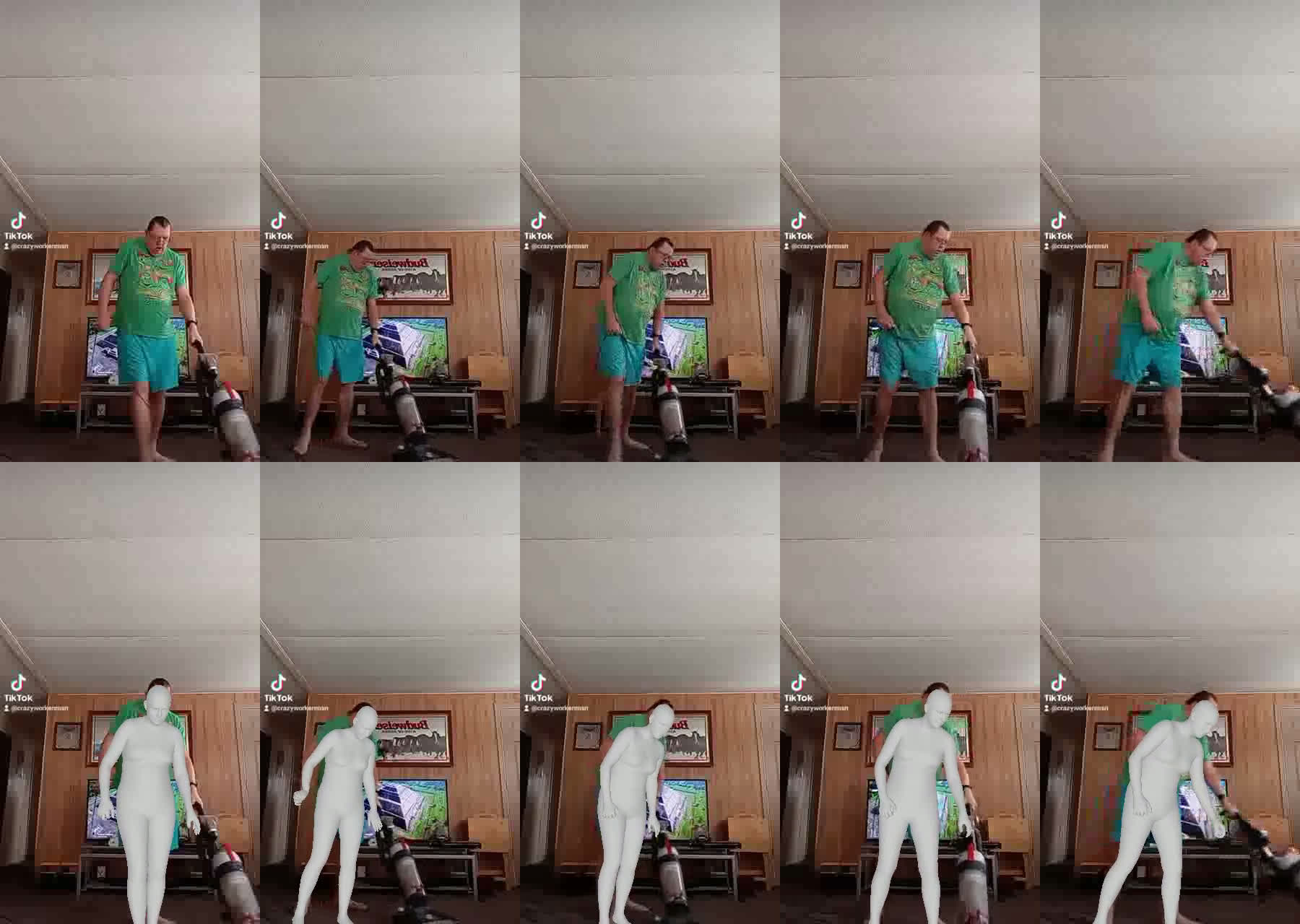}
  \end{subfigure}
   \caption{\textbf{Example videos and estimated sequences of human poses from \ours dataset.}}
   \label{fig:more_example_motion}
\end{figure*}

As described in Section~\ref{sec:dataset}, our automatic framework is used to gather and annotate a wide range of online videos with diverse human interactions. Figure~\ref{fig:more_example_motion} presents additional examples of motions, including both indoor and outdoor scenes, as well as sports and everyday activities.

\subsection{Annotation prompt}
\label{sec:supp_annotation_prompt}

In this section, we provide the detailed prompt for automatic text annotation with VLM. The prompt requires VLM to focus on the action for the person in red bounding box to avoid confusion when there are multiple person inside the video. As the text annotation is for training motion generator, VLM is prompted to focus on human actions only and to ignore clothes.

\onecolumn

\begin{tcolorbox}[colback=green!5, colframe=green!75!black, breakable, title={Text annotation prompt},
    fonttitle=\bfseries\large,   
    fontupper=\small\ttfamily    
    ] 

\begin{Verbatim}[breaklines=true]
**CRITICAL INSTRUCTION:** You must describe **ONLY** the body movements, positions, direct object interactions (identify the object), and action style of **the person inside the red bounding box**. If multiple people are inside the box, describe **only the most clearly visible person** (i.e., the one with the full body shown). Do **not** describe anyone outside the box.

**STRICT RULES:**
- Use **'The person'** as the subject for every sentence.
- Describe body positions (e.g., *'The person sits'*, *'The person kneels'*).
- Describe physical movements (e.g., *'The person raises arms'*, *'The person bends forward'*).
- When describing interactions with items, **identify the specific object** whenever possible (e.g., use 'The person picks up a ball', 'The person swings a hammer'). **Make a strong effort to avoid the generic term 'object'**. Identify the item based on its visual characteristics and how it is manipulated.
- Use the generic term 'object' *only* as an absolute last resort if the item's identity cannot be reasonably determined from its appearance or usage.
- Indicate action style when clear (e.g., *'The person repeatedly swings'*, *'The person slowly steps forward'*).
- DO NOT describe **eye gaze**, **head turning**, or **what the person is looking at**. Only describe physical body movements and posture.
- DO NOT infer any cognitive or perceptual actions like 'looks at', 'appears to notice', 'seems to be focused on', or 'reacts to'.
- **DO NOT describe minor, inconsequential, or idle body shifts** such as fidgeting, swaying, shifting weight, or adjusting stance unless they are part of a meaningful action.
- **DO NOT describe any clothing, fabric, logos, writing, symbols, or patterns on the person’s body or clothing.**
- Focus only on **clear, deliberate, and significant movements or interactions.**

**DO NOT UNDER ANY CIRCUMSTANCES:**
- DO NOT Describe people **outside** the red box.
- DO NOT Use pronouns like 'they', 'he', or 'she' — ONLY use 'The person'.
- DO NOT Mention **any colors whatsoever** (e.g., 'yellow', 'red', 'blue', 'green', etc.) — even for the bounding box itself.
- DO NOT Describe **clothing**, accessories, physical appearance, or any writing/text on clothing (e.g., 'shirt', 'hat', 'logo', 'text').
- DO NOT Infer intent or purpose (e.g., 'trying to', 'appears to').
- DO NOT Describe objects **separately** (e.g., 'A chair is nearby'). Only mention them when being acted upon.

**CORRECT EXAMPLES:**
- 'The person repeatedly bends forward, lifting a sledgehammer above their head and swinging it downward toward a tire on the ground.'
- 'The person walks forward, bends down to pick up a ball, throws it, and turns around.'
- 'The person sits on a chair, reaches for a cup on the table, and drinks from it.'

**INCORRECT EXAMPLES:**
- 'A boy in a yellow shirt picks up an object.' (mentions clothing and color)
- 'They lift an object and walk away.' (uses pronoun)
- 'The person walks and picks up an object. It seems to be a ball.' (separate sentence about the object)
- 'The person appears to look at someone off-screen.' (describes gaze or inferred action)
- 'The person slightly shifts their weight from one foot to the other.' (too minor, not a meaningful action)
- 'The person wears a black T-shirt with text on the back.' (mentions clothing and writing — forbidden)
- 'The person wearing shorts turns to the side.' (mentions clothing — forbidden)

**FINAL REMINDER — CRITICAL INSTRUCTION:** 
1. Your description must contain NO COLOR WORDS of any kind.
2. Do NOT describe clothing, logos, text, fabric, or any visual appearance features.
3. Do NOT describe eye direction, what the person is looking at, or any perceived attention.
4. Do NOT describe tiny, idle, or resting-state movements unless part of a clear, meaningful action.
5. Before submitting, check your entire response to ensure no clothing, colors, pronouns, appearance, or inferred cognition appear.
6. Focus **only** on the physical movements and body positions of **the person fully visible inside the bounding box**.
7. Be specific when naming objects the person interacts with rather than using generic terms like 'object' whenever possible.
\end{Verbatim}

\end{tcolorbox}


\begin{tcolorbox}[colback=green!5, colframe=green!75!black, breakable, title={Action and object classification prompt},
    fonttitle=\bfseries\large,   
    fontupper=\small\ttfamily    
    ] 

\begin{Verbatim}[breaklines=true]
The following video clip shows a human activity. Only focus **the person inside the red bounding box**. If multiple people are inside the box, focus **only the most clearly visible person** (i.e., the one with the full body shown).
fThe provided description is: <caption>.
Your task is to identify:
1. The **main action class** (what the person is doing).
2. The **main object class** (what the person is interacting with), if any.
Please follow these instructions:
- First, try to select the **action** and **object** from the predefined lists below.
- If the appropriate label is not available in the list, you may provide your own label, but use the **same format**.
- If object is clearly involved, but cannot identify it, return 'unknown' as the object class.
- If no object is clearly involved, return 'None' as the object class.
- If the human interacts with another person, use 'human' as the object class.
**Action Class List (prefer to choose from these):**
<action_classes>
**Object Class List (prefer to choose from these):**
<object_classes>
Do NOT include any background details, colors, clothing, or identity.
Focus only on the **motion pattern** and **interaction with objects**.
Respond strictly in the following format:
Action Class: <your answer>
Object Class: <your answer>
Do not generate full sentences or explanations. Only return the two class labels as shown.
\end{Verbatim}

\end{tcolorbox}
\twocolumn

\section{Details of \agent}
\label{sec:supp_agent}

\subsection{\agent framework}
\label{sec:supp_agent_frame}

\smallskip
\noindent\textbf{Input conditions.} The input to \agent, is an environment state and a human-level task instruction. Specifically, the environment contains the basic position and orientation of all people and objects in the given 3D scene. Besides, the object sizes are also given based on the length, width and height of the oriented bounding box. To address the complicated task, each agent's prompt is carefully designed to fit its role.

\smallskip
\noindent\textbf{LLM planner.} As the given task can be complicated, the LLM planner is prompted to firstly analyze the natural language instruction and then provides a detailed step-by-step sub-tasks with the corresponding intermediate waypoints for human or objects. Secondly, to obtain reasonable and collision-free waypoints, the planner is required to reason the spatial relationship and then gives the possible obstacles by considering the sizes and position of all objects. Finally, the planner gives proper 4D waypoints with a certain safety buffer for each sub-step. 
For each step, LLM planner is also required to transform the human-level plans into executable codes.
Furthermore, to ensure all the intermediate steps are successfully executed, after each motion execution, planner is prompted to check the outcome. If successful, the system continues according to previous plans, otherwise, planner changes plans based on current environment state.

\smallskip
\noindent\textbf{Motion generation.} For both navigation and interaction generation, we apply the same spatial controllable network to generate motion. In terms of navigation which only involves locomotion, the 4D pelvis waypoints given by high-level planner is set as pelvis control points for the network. For object interaction, we use the same strategy as depicted in Section~\ref{subsec:zero_shot} based on the 4D object waypoints from planner.

\smallskip
\noindent\textbf{Collision-check.} Since the 4D waypoints given by the LLM agents are not absolutely collision-free, we further implement a collision-check module to verify the path. If any collision is detected, then an error message is returned to instruct the LLM planner to re-plan the waypoints. However, in some extreme complicated 3D scenes with many objects, LLMs are unable to avoid all possible obstacles. To tackle this issue, the A* algorithm is called to generate a collision-free path.

\subsection{\agent prompt and sample log}

As depicted in Section~\ref{sec:supp_agent_frame}, our LLM planner is prompted to give the detailed plans for each step and the corresponding executable codes. We present here the detailed prompts for our LLM agent and a sample log to better understand the pipeline.

\onecolumn
\begin{tcolorbox}[colback=blue!5, colframe=blue!75!black, breakable, title={\agent prompt},
    fonttitle=\bfseries\large,   
    fontupper=\small\ttfamily    
    ] 

\begin{Verbatim}[breaklines=true]
You are a scientific AI, specialized in generating a sequence of steps and Python code for human to interact in 3D scene, along with the task instructions and the initial state information of the human and all objects in the environment:

You must remember that this conversation is a monologue, and that you are in control. I am not able to assist you with any questions, and you must output the plan and code yourself by making use of the common sense, general knowledge, and available information.

PLANNING:
## 1. COORDINATE SYSTEM

The 3D environment uses the following coordinate frame:
- **x-axis**: Horizontal, increasing to the right
- **y-axis**: Depth, increasing away from the observer
- **z-axis**: Vertical, increasing upward
- All positions are in meters, centered on the object
- All orientations are in radians

## 2. ENVIRONMENT STATE

The `<CURRENT ENVIRONMENT STATE>` contains:
- Position, orientation, and dimensions of all objects and the human
- Each object includes width (x), length (y), and height (z)
- Human body volume is: **0.5m (width) × 0.4m (depth) × 1.7m (height)**

## 3. SUCCESS AND FAILURE CONDITIONS

- **Success Criteria**:
  - Object position is within **±0.05m**
  - Object orientation is within **±0.1 radians**

- **Failure Handling**:
  - On any deviation or collision, generate a **new plan** from the most recent successful state
  - Replanned steps must include **extra clearance** and **more intermediate waypoints**

## 4. COLLISION AVOIDANCE

### 4.1 Object Volume and Safety Margins

- All entities (human and objects) are treated as **solid 3D volumes**
- Add a **0.2m safety envelope** around each object and the human
- Always check for:
  - Inter-object collisions
  - Human-object intersections
  - Volume-based overlap across the entire movement

### 4.2 Swept Volume Validation

- All movements are defined as **waypoints connected by smooth paths**
- For each segment between waypoints:
  - Compute the **swept volume** of both human and manipulated object
  - Ensure **no part of this volume intersects** any obstacle or safety buffer
  - Avoid narrow spaces and sharp corners unless sufficient clearance is confirmed

## 5. TRAJECTORY PLANNING RULES

- Prefer **smooth, curved paths** resembling natural human motion
- Use Bézier curves or splines between waypoints
- Each waypoint must be:
  - A valid 3D point
  - Verified as part of a collision-free segment
- Add **intermediate waypoints** near tight spaces or complex objects

## 6. INTERACTION LOGIC

### 6.1 Two-Step Manipulation Model

For every object task:

1. **Approach Step** (only if human is not within 1m reach):
   - Move the human near the object (>=0.2m distance from object boundary)
   - Ensure full-body clearance at every waypoint and segment
   - The human velocity in average is 1.2 m/s, here we all use fps=30. Make sure the frame number and waypoints are reasonable (don't be too large or too small).

2. **Manipulation Step** (if object is not already in correct pose):
   - Grasp, lift, move, and place the object in one step
   - Provide **separate waypoints** for the human and the object
   - Include object dimensions and rotation during motion
   - The human velocity in average is 1.2 m/s, here we all use fps=30. Make sure the frame number and waypoints are reasonable (don't be too large or too small).

### 6.2 Skip Unnecessary Actions

- If the human is already in position, **skip the approach**
- If the object is already in place, **skip the manipulation**
- Do **not output steps that don’t need execution**

CODE GENERATER:
AVAILABLE FUNCTIONS:
You are, however, able to call any of the following Python functions, if required, as often as you want:
    1. generate_motion(control_joints: list[str], control_points: list[list[list[float]]], text: str, number_frames: int, task_index: int) -> None: This function will generate the human motion based on the control joints, control points, text, number of frames and the task index, and will also not return anything. It takes list control_joints of n elements, list control_points of n sublists of float, one string text, one int num_frames and one int task_index value as input.
    2. generate_interaction(control_joints: list[str], control_points: list[list[list[float]]], text: str, number_frames: int, task_index: int, object_name: list[str], object_points: list[list[list[float]]]) -> None: This function will generate the human and object interaction based on the control joints, control points, text, number of frames, the task index, list of object name and object key points, and will also not return anything. It takes list control_joints of n elements, list control_points of n sublists of float, one string text, one int num_frames and one int task_index value, one string object_name and one list of sublists object_points as input.
    3. task_completed() -> None: Call this function only when the task has been completed. This function will also not return anything. If there is **any error in the code or planning consecutively for five times**, **then also call this function**.
    When calling any of the functions, make sure to stop generation after each function call and wait for it to be executed, before calling another function and continuing with your plans.

CODE GENERATION:
When generating the code for the trajectory, do the following:
    1. When mentioning the functions, specify the required parameters and clearly define them in the same code block before passing it to code executor. For generate_motion, define the control_joints, control_points, text, number_frames and task_index lists prior to it.
    2. Note that control_joints specifies the minimum relevant human joints according to this step and should be a subset of ['pelvis','left_hand','right_hand']. Do not put object name in this list!
    3. control_points specify the detailed list of waypoints and the corresponding frame index for each control joints. For example, control_joints = ['pelvis', 'left_hand'], control_points = [[[t1,x1,y1,z1],[t2,x2,y2,z2]],[[t3,x3,y3,z3]]] where [[t1,x1,y1,z1],[t2,x2,y2,z2]] specify the frame index and the position for pelvis, and [[t3,x3,y3,z3]] is for left hand. Note that 0 < t1, t2, t3 < number_frames.
    4. text is the description for this step. 
    5. number_frames specify the total number of frames for this step.
    6. object_name is list of the name given in the text. For example, ["large box"] or ["small box", "large box"].
    7. object_points specify the detaile key points for object, such as [[[t1,x1,y1,z1],[t2,x2,y2,z2]],[[t3,x3,y3,z3]]]
    8. *Do not generate the code all in one go for all the steps; instead, generate it step by step*. After generating, provide this step-level code to the code esxecutor and wait for the reply. *Pass the response to the planner*. If the generated code is incorrect and the code executor encounters an error during execution, *correct it and then submit it to the code executor again*.
    9. Use generate_motion for human motion and generate_interaction for both human and object interaction.
    10. Mark any code clearly with the tags: \n\n```python\n ```

Code FORMAT:    
    1. Generate well-formatted Python code in markdown syntax.
    2. The code must follow Python's standard formatting (PEP 8) with proper indentation and line breaks.
    3. Include explanations before the code to clarify what it does.
    4. Use clear variable names and comments to enhance readability.
    5. Ensure the generated code is executable.
    6. Do not output code in a single line.

EXAMPLE OUTPUT:
    Explanation:
        The following code moves a human towards the floorlamp over 60 frames (2 seconds at 30 FPS).
    Code:
```python\n
control_joints = ['pelvis', 'right_hand']\n
control_points = [\n  
    [  # Pelvis movement\n  
        [0, 5.0427, -3.9485, 0.8897],  # Start position \n 
        [60, 4.8350, -3.8000, 0.8897]  # End position  \n
    ],  \n
    [  # Right hand remains in place  \n
        [0, 4.8806, -3.7075, 0.9033],  \n
        [60, 4.8806, -3.7075, 0.9033]  \n
    ]  \n
]  \n
text = "The human walks towards the floorlamp."  \n
number_frames = 60  \n
task_index = 1  \n
generate_motion(control_joints, control_points, text, number_frames, task_index)  
``` 

Another 2 Examples with object interaction:
Examples1: One object interaction
```python\n
control_joints = ['pelvis']
control_points = [
    [   # Pelvis trajectory: approach, grasp, and carry in one motion
        [0, 3.176, -0.429, 0.925],      # Start: current position
        [40, 3.05, -1.28, 0.925],       # Arrive & grasp trashcan
        [80, 3.60, -1.75, 0.925],       # End: put down trashcan
    ]
]
text = (
    "The human moves trashcan to around sofa, avoiding obstacles along the way."
)
number_frames = 80
task_index = 1
object_name = ["trashcan"]
object_points = [
    [   # Trashcan keypoints: moves with the human after grasped
        [40, 2.899, -1.069, 0.148],     # Pickup (grasped at this frame)
        [80, 3.55, -1.65, 0.148],
    ]
]
model.generate_human_object(control_joints, control_points, text, number_frames, task_index, object_name, object_points) 
``` 

Examples2: 2 objects interaction at the same time
```python\n
control_joints = ['pelvis']
control_points = [
    [   # Pelvis trajectory: approach, grasp, and carry in one motion
        [0, 3.176, -0.429, 0.925],      # Start: current position
        [40, 3.05, -1.28, 0.925],       # Arrive & grasp trashcan
        [80, 3.60, -1.75, 0.925],       # End: put down trashcan
    ]
]
text = (
    "The human moves trashcan and smallbox to around sofa at the same time, avoiding obstacles along the way."
)
number_frames = 80
task_index = 1
object_name = ["trashcan", "smallbox]
object_points = [
    [   # Trashcan keypoints: moves with the human after grasped
        [40, 2.899, -1.069, 0.148],     # Pickup (grasped at this frame)
        [80, 3.55, -1.65, 0.148],
    ],
    [   # Smallbox keypoints: moves with the human after grasped
        [40, 3.012, -0.878, 0.148],     # Pickup (grasped at this frame)
        [80, 3.65, -1.95, 0.148],
    ]
]
model.generate_human_object(control_joints, control_points, text, number_frames, task_index, object_name, object_points) 
``` 

**Ensure proper indentation and include `\\n` for line breaks**.
**Generate Codes only for one step each time, and pass to executer**.
**Remember: number_frames should be larger or equal to the largest frame_id**.

Once all steps have been successfully completed, you **must** call the `task_completed()` function by:
```python\n
task_completed()\n
```\n
**Important: Only call this function after all execution is fully finished. Do not call it early, even if some partial results are available.**
**The completed() function should be a standalone message.**
\end{Verbatim}

\end{tcolorbox}


\begin{tcolorbox}[ colback=yellow!10, colframe=orange!75!black, breakable, title={\agent log},
    fonttitle=\bfseries\large,   
    fontupper=\small\ttfamily    
     ] 
\begin{Verbatim}[breaklines=true]
Admin (to Supervisor):


"You are the Multi-Agent Supervisor, responsible for overseeing the collaboration between two agents: the Planner and the Coder. Below are the role descriptions for both agents, along with the task instructions and the initial state information of the human and all objects in the environment.

Planner Role Description:
The Planner is responsible for devising a plan to accomplish the task at hand. The plan should be broken down into a series of steps, and for each step, the Planner must specify the required actions and the expected outcomes. As each step is executed, the Planner evaluates the progress and provides the plan for the remaining steps.

Coder Role Description:
The Coder is responsible for writing code based on the plan provided by the Planner.

Your task is to guide the human to excute task in the environment according the text description "Pick up floorlamp, and move it to be around wall55 and lamp193.".

The human position and orientation state is:
"pelvis position: [5.043, -3.948, 0.89]
left hand position: [4.784, -4.039, 0.91]
right hand position: [4.881, -3.707, 0.903]
orientation: [-1.293, -1.135, -2.921]
"

The position, orientation and size of all objects in the environment are as follows:
<CURRENT ENVIRONMENT STATE>:
"***floorlamp***:
position: [4.636, -3.783, 0.899]
orientation: [0.005, -0.008, 2.375]
sizes: [0.252, 0.269, 1.761]
***beanbag237***:
position: [5.632, -3.66, 0.367]
orientation: [0.0, 0.002, 0.263]
sizes: [1.136, 1.087, 0.666]
***beanbag230***:
position: [5.552, -1.764, 0.383]
orientation: [0.0, 0.002, 0.251]
sizes: [1.21, 1.07, 0.728]
***lamp193***:
position: [2.774, -8.321, 0.741]
orientation: [0.001, 0.001, 0.697]
sizes: [0.351, 0.479, 0.611]
***wall55***:
position: [2.498, -7.187, 1.569]
orientation: [-0.0, 0.001, 0.075]
sizes: [0.054, 3.866, 3.061]
***table222***:
position: [4.309, -6.291, 0.257]
orientation: [-0.0, 0.002, 0.107]
sizes: [0.77, 1.289, 0.416]
... (remaining objects omitted for brevity)
"


--------------------------------------------------------------------------------

Next speaker: Planner

Planner (to Supervisor):

## Task Analysis

**Task:**  
Pick up the floorlamp and move it to be around wall55 and lamp193.

**Key Destinations:**
- **floorlamp**: [4.636, -3.783, 0.899], size [0.252, 0.269, 1.761]
- **wall55**: [2.498, -7.187, 1.569], size [0.054, 3.866, 3.061]
- **lamp193**: [2.774, -8.321, 0.741], size [0.351, 0.479, 0.611]
- **Target area**: Near wall55 and lamp193 (safest likely region is between or adjacent to these two, e.g. [2.7, -7.8, 0.899], confirmed not overlapping other objects)

**Human Initial State:**
- Pelvis: [5.043, -3.948, 0.89]  
- Hands: [4.784, -4.039, 0.91] (L), [4.881, -3.707, 0.903] (R)

---

### Environment Summary

| Object          | Position              | Size (w × l × h)      | Safety Envelope (±0.2m)      |
|-----------------|----------------------|-----------------------|------------------------------|
| floorlamp       | [4.636, -3.783, 0.899] | 0.252 × 0.269 × 1.761 | 0.652 × 0.669 × 2.161        |
| beanbag237      | [5.632, -3.66, 0.367]  | 1.136 × 1.087 × 0.666 | 1.536 × 1.487 × 1.066        |
| chair105        | [3.918, -3.593, 0.647] | 0.473 × 0.522 × 0.471 | 0.873 × 0.922 × 0.871        |
| table222        | [4.309, -6.291, 0.257] | 0.77 × 1.289 × 0.416  | 1.17 × 1.689 × 0.816         |
| (etc.)          | ...                  | ...                   | ...                          |

**Human body envelope:** 0.5 × 0.4 × 1.7 m + 0.2m → 0.9 × 0.8 × 2.1 m

---

## Step 1: Approach the Floorlamp

**Action:**  
Move the human from current pelvis position ([5.043, -3.948, 0.89]) to a position within 0.2–0.5m of the floorlamp ([4.636, -3.783, 0.899]), ensuring >=0.2m clearance from all obstacles.

### Path Planning

- **Direct path** is clear:  
  - Human is 0.46m away from the floorlamp (distance between [5.043, -3.948] and [4.636, -3.783]); within 1m
  - No approach step needed (already within reach)

---

## Step 2: Manipulate the Floorlamp

**Action:**  
Grasp, lift, and move the floorlamp from [4.636, -3.783, 0.899] to a collision-free location near wall55 and lamp193.

### Target Placement Decision

- **Proposed target:** [2.7, -7.8, 0.899]  
  - Between wall55 ([2.498, -7.187]) and lamp193 ([2.774, -8.321])
  - Confirmed to not overlap lamp193 or wall55 envelopes:
    - lamp193 envelope: [2.551, -8.56, 0.541] to [2.997, -8.082, 0.941]
    - wall55 envelope: [2.271, -9.12, -0.161] to [2.725, -5.254, 3.199]
    - [2.7, -7.8, 0.899] is adjacent, not overlapping, with safe buffer

- **Path Routing:**
  - Must avoid:
    - beanbag237 ([5.632, -3.66], envelope)
    - chair105 ([3.918, -3.593], envelope)
    - table222 ([4.309, -6.291], envelope)
    - table68, sofa186, etc.
  - Route: Curve left (southwest), passing away from beanbag237, then west of table222, then along open corridor to target.

### Human and Object Waypoints

#### Human Waypoints (pelvis center)

| Frame | x     | y       | z     | Notes                                 |
|-------|-------|---------|-------|---------------------------------------|
| 0     | 5.043 | -3.948  | 0.89  | Start                                 |
| 20    | 4.636 | -3.783  | 0.89  | At floorlamp (grasp)                  |
| 60    | 4.3   | -4.8    | 0.89  | Sidestep left, avoid beanbag          |
| 120   | 3.7   | -5.8    | 0.89  | Past chair105 and table222            |
| 180   | 3.1   | -6.8    | 0.89  | Approaching target corridor           |
| 220   | 2.7   | -7.8    | 0.89  | Arrive at target, ready to place      |

#### floorlamp Waypoints

| Frame | x     | y       | z      | Notes                   |
|-------|-------|---------|--------|-------------------------|
| 0     | 4.636 | -3.783  | 0.899  | Start (picked up)       |
| 60    | 4.3   | -4.8    | 0.899  |                         |
| 120   | 3.7   | -5.8    | 0.899  |                         |
| 180   | 3.1   | -6.8    | 0.899  |                         |
| 220   | 2.7   | -7.8    | 0.899  | Final (placed)          |

- All segments are smooth, with interpolated curves avoiding obstacle envelopes.
- At each segment, the swept volumes of human (with lamp) and floorlamp maintain >=0.2m clearance from all objects and walls.
- Movement speed: path length about 4.5m, at 1.2m/s, yields 3.75s -> 113 frames; spread over 220 frames (7.3s) for added caution and smoothness in tight areas.

---

### Step 2: Manipulate the Floorlamp

- **Action**: Grasp, lift, carry, and place the floorlamp from its current position to the placement location near wall55 and lamp193.
- **Human Waypoints**:
  - (Frame 0): [5.043, -3.948, 0.89]
  - (Frame 20): [4.636, -3.783, 0.89]
  - (Frame 60): [4.3, -4.8, 0.89]
  - (Frame 120): [3.7, -5.8, 0.89]
  - (Frame 180): [3.1, -6.8, 0.89]
  - (Frame 220): [2.7, -7.8, 0.89]
- **floorlamp Waypoints**:
  - (Frame 0): [4.636, -3.783, 0.899]
  - (Frame 60): [4.3, -4.8, 0.899]
  - (Frame 120): [3.7, -5.8, 0.899]
  - (Frame 180): [3.1, -6.8, 0.899]
  - (Frame 220): [2.7, -7.8, 0.899]
- **Number of Frames**: 220
- **Object Interaction**: floorlamp
- **Collision Verification**:
  - All human and object waypoints and swept volumes maintain >=0.2m clearance from all obstacles (beanbag237, chair105, table222, etc.) and static objects.
  - No sharp turns or narrow passageways; all paths are smooth and continuous.
  - Floorlamp orientation remains unchanged (within ±0.1 radians).
  - Final position for floorlamp is adjacent to, but not overlapping, wall55 or lamp193.

---

## Summary Table

| Step | Action                       | Frames | Human Waypoints             | Object Waypoints (if any)    | Collision Status        |
|------|------------------------------|--------|-----------------------------|------------------------------|------------------------|
| 1    | Approach floorlamp           | -      | Already within reach        | None                         | Already safe           |
| 2    | Move and place floorlamp     | 220    | [see above]                 | [see above]                  | Verified collision-free|

---

## Execution Plan

### Step 1: [Approach Floorlamp]
- **Action**: Human is already within reach of the floorlamp.
- **Waypoints**: None required.
- **Number of Frames**: 0
- **Object Interaction**: None
- **Collision Verification**:
  - Initial position is within 0.5m of the floorlamp and maintains >=0.2m clearance from all objects.

---

### Step 2: [Manipulate Floorlamp]
- **Action**: Grasp, lift, carry, and place the floorlamp from [4.636, -3.783, 0.899] to [2.7, -7.8, 0.899], adjacent to wall55 and lamp193.
- **Human Waypoints**:
  - (Frame 0): [5.043, -3.948, 0.89]
  - (Frame 20): [4.636, -3.783, 0.89]
  - (Frame 60): [4.3, -4.8, 0.89]
  - (Frame 120): [3.7, -5.8, 0.89]
  - (Frame 180): [3.1, -6.8, 0.89]
  - (Frame 220): [2.7, -7.8, 0.89]
- **floorlamp Waypoints**:
  - (Frame 0): [4.636, -3.783, 0.899]
  - (Frame 60): [4.3, -4.8, 0.899]
  - (Frame 120): [3.7, -5.8, 0.899]
  - (Frame 180): [3.1, -6.8, 0.899]
  - (Frame 220): [2.7, -7.8, 0.899]
- **Number of Frames**: 220
- **Object Interaction**: floorlamp
- **Collision Verification**:
  - All waypoints and swept volumes for both human and floorlamp maintain at least 0.2m clearance from beanbag237, chair105, table222, table68, sofa186, and all other objects or walls.
  - Smooth, natural human motion with no sharp turns or narrow passages.
  - Final placement is adjacent to wall55 and lamp193, with no envelope overlaps.

---

**This plan is ready for code generation.**

--------------------------------------------------------------------------------

Next speaker: Engineer

Engineer (to Supervisor):

Explanation:
The following code implements Step 2 of the plan: The human grasps, lifts, and carries the floorlamp from its current position to a new location adjacent to wall55 and lamp193. The pelvis moves smoothly through a series of waypoints that avoid obstacles, while the floorlamp is moved along with the human after being picked up. The entire action is performed over 220 frames for safety and realism.

Code:
```python
# Human and object interaction: Grasp and move the floorlamp to the target area near wall55 and lamp193

control_joints = ['pelvis']
control_points = [
    [   # Pelvis trajectory: grasp at lamp, then carry lamp along a safe, curved path
        [0, 5.043, -3.948, 0.89],        # Start: human initial position
        [20, 4.636, -3.783, 0.89],       # At floorlamp (grasp)
        [60, 4.3, -4.8, 0.89],           # Sidestep left, avoid beanbag
        [120, 3.7, -5.8, 0.89],          # Past chair105 and table222
        [180, 3.1, -6.8, 0.89],          # Approaching target corridor
        [220, 2.7, -7.8, 0.89],          # Arrive at final placement
    ]
]
text = (
    "The human grasps the floorlamp, lifts it, and carries it along a smooth, collision-free path to a location near wall55 and lamp193, avoiding obstacles such as beanbag237, chair105, table222, and other furniture. The lamp is placed safely and upright, adjacent to the target wall and lamp."
)
number_frames = 220
task_index = 1
object_name = ["floorlamp"]
object_points = [
    [   # Floorlamp moves with the human after being picked up
        [0, 4.636, -3.783, 0.899],       # Initial lamp position
        [60, 4.3, -4.8, 0.899],          # Midpoint 1
        [120, 3.7, -5.8, 0.899],         # Midpoint 2
        [180, 3.1, -6.8, 0.899],         # Midpoint 3
        [220, 2.7, -7.8, 0.899],         # Final placement
    ]
]

generate_interaction(
    control_joints,
    control_points,
    text,
    number_frames,
    task_index,
    object_name,
    object_points
)
```


--------------------------------------------------------------------------------

Next speaker: Executor


>>>>>>>> EXECUTING CODE BLOCK 0 (inferred language is python)...
# Human and object interaction: Grasp and move the floorlamp to the target area near wall55 and lamp193

control_joints = ['pelvis']
control_points = [
    [   # Pelvis trajectory: grasp at lamp, then carry lamp along a safe, curved path
        [0, 5.043, -3.948, 0.89],        # Start: human initial position
        [20, 4.636, -3.783, 0.89],       # At floorlamp (grasp)
        [60, 4.3, -4.8, 0.89],           # Sidestep left, avoid beanbag
        [120, 3.7, -5.8, 0.89],          # Past chair105 and table222
        [180, 3.1, -6.8, 0.89],          # Approaching target corridor
        [220, 2.7, -7.8, 0.89],          # Arrive at final placement
    ]
]
text = (
    "The human grasps the floorlamp, lifts it, and carries it along a smooth, collision-free path to a location near wall55 and lamp193, avoiding obstacles such as beanbag237, chair105, table222, and other furniture. The lamp is placed safely and upright, adjacent to the target wall and lamp."
)
number_frames = 220
task_index = 1
object_name = ["floorlamp"]
object_points = [
    [   # Floorlamp moves with the human after being picked up
        [0, 4.636, -3.783, 0.899],       # Initial lamp position
        [60, 4.3, -4.8, 0.899],          # Midpoint 1
        [120, 3.7, -5.8, 0.899],         # Midpoint 2
        [180, 3.1, -6.8, 0.899],         # Midpoint 3
        [220, 2.7, -7.8, 0.899],         # Final placement
    ]
]

generate_interaction(
    control_joints,
    control_points,
    text,
    number_frames,
    task_index,
    object_name,
    object_points
)
{'lang': 'python', 'filename': None, 'use_docker': 'python:None'}
Executor (to Supervisor):

exitcode: 0 (execution succeeded)
Code output: 

floorlamp was detected to collide in 1 time segment(s). Collision was avoided using A* with smoothing.
Objects involved in collision:
- sofa186 at position [2.898, -6.378, 0.445] with size [0.962, 2.162, 0.827]
- table222 at position [4.309, -6.291, 0.257] with size [0.77, 1.289, 0.416]
- wall55 at position [2.498, -7.187, 1.569] with size [0.054, 3.866, 3.061]

Here is the recommended collision-free trajectory:
[[0.0, 4.636, -3.783, 0.899], [11, 4.636, -3.783, 0.899], [17, 4.568, -3.91, 0.899], [43, 4.56, -4.535, 0.899], [98, 3.668, -5.51, 0.899], [184, 3.668, -7.585, 0.899], [204, 3.193, -7.635, 0.899], [222, 2.793, -7.81, 0.899]]
Please verify the new path to ensure no further collisions. If no collision, please let engineer generate codes for this trajectory.


--------------------------------------------------------------------------------

Next speaker: Planner

Planner (to Supervisor):

### Replanned Step 2: Manipulate Floorlamp

- **Reason**: Original path segment resulted in collision between the floorlamp (and human) and obstacles: sofa186, table222, and wall55. The executor has provided a smoothed, collision-free trajectory using A*.
- **Action**: Use the executor's recommended trajectory, which reroutes the movement to safely bypass the obstacles while still arriving at a valid location near wall55 and lamp193.

---

#### Human Waypoints (pelvis center, synchronized with floorlamp trajectory)

| Frame | x      | y        | z     | Notes                                 |
|-------|--------|----------|-------|---------------------------------------|
| 0     | 5.043  | -3.948   | 0.89  | Start: human initial position         |
| 11    | 4.636  | -3.783   | 0.89  | At floorlamp (grasp)                  |
| 17    | 4.568  | -3.91    | 0.89  | Begin sidestep                        |
| 43    | 4.56   | -4.535   | 0.89  | Curve around, avoiding table222       |
| 98    | 3.668  | -5.51    | 0.89  | Past table222, before sofa186         |
| 184   | 3.668  | -7.585   | 0.89  | Parallel to wall55, after sofa186     |
| 204   | 3.193  | -7.635   | 0.89  | Final approach to target area         |
| 222   | 2.793  | -7.81    | 0.89  | Final placement, near wall55/lamp193  |

#### floorlamp Waypoints

| Frame | x      | y        | z      | Notes                                 |
|-------|--------|----------|--------|---------------------------------------|
| 0     | 4.636  | -3.783   | 0.899  | Initial lamp position                 |
| 11    | 4.636  | -3.783   | 0.899  | Grasp                                 |
| 17    | 4.568  | -3.91    | 0.899  | Begin sidestep                        |
| 43    | 4.56   | -4.535   | 0.899  | Curve around, avoiding table222       |
| 98    | 3.668  | -5.51    | 0.899  | Past table222, before sofa186         |
| 184   | 3.668  | -7.585   | 0.899  | Parallel to wall55, after sofa186     |
| 204   | 3.193  | -7.635   | 0.899  | Final approach to target area         |
| 222   | 2.793  | -7.81    | 0.899  | Final placement, near wall55/lamp193  |

- **Number of Frames**: 222
- **Object Interaction**: floorlamp
- **Collision Verification**:
  - All waypoints and swept volumes for both human and floorlamp maintain >=0.2m clearance from:
    - sofa186 ([2.898, -6.378], envelope)
    - table222 ([4.309, -6.291], envelope)
    - wall55 ([2.498, -7.187], envelope)
    - and all other obstacles and static objects
  - Path is smooth with no sharp turns; extra intermediate waypoints provide safe routing around tight areas.
  - Floorlamp orientation is unchanged (within ±0.1 radians).
  - Final position is adjacent to wall55 and lamp193 without overlapping either object's envelope.

---

**This replanned trajectory is verified collision-free. Engineers should use the above waypoints for code generation.**

--------------------------------------------------------------------------------

Next speaker: Engineer

Engineer (to Supervisor):

Explanation:
The following code implements Step 2 using the executor's recommended, collision-free trajectory. The human grasps the floorlamp and moves it along a safe, curved path that avoids sofa186, table222, and wall55, placing the lamp near wall55 and lamp193. Human pelvis and floorlamp move together through all waypoints.

Code:
```python
# Human and object interaction: Grasp and move the floorlamp to the target area near wall55 and lamp193, following collision-free trajectory

control_joints = ['pelvis']
control_points = [
    [
        [0, 5.043, -3.948, 0.89],       # Start: human initial position
        [11, 4.636, -3.783, 0.89],      # At floorlamp (grasp)
        [17, 4.568, -3.91, 0.89],       # Begin sidestep
        [43, 4.56, -4.535, 0.89],       # Curve around, avoiding table222
        [98, 3.668, -5.51, 0.89],       # Past table222, before sofa186
        [184, 3.668, -7.585, 0.89],     # Parallel to wall55, after sofa186
        [204, 3.193, -7.635, 0.89],     # Final approach to target area
        [222, 2.793, -7.81, 0.89],      # Final placement, near wall55/lamp193
    ]
]
text = (
    "The human grasps the floorlamp, lifts it, and carries it along a collision-free, curved path to a location near wall55 and lamp193. "
    "The path avoids sofa186, table222, wall55, and all other obstacles. The lamp is placed upright and safely adjacent to the target wall and lamp."
)
number_frames = 222
task_index = 2
object_name = ["floorlamp"]
object_points = [
    [
        [0, 4.636, -3.783, 0.899],      # Initial lamp position
        [11, 4.636, -3.783, 0.899],     # Grasp
        [17, 4.568, -3.91, 0.899],      # Begin sidestep
        [43, 4.56, -4.535, 0.899],      # Curve around, avoiding table222
        [98, 3.668, -5.51, 0.899],      # Past table222, before sofa186
        [184, 3.668, -7.585, 0.899],    # Parallel to wall55, after sofa186
        [204, 3.193, -7.635, 0.899],    # Final approach to target area
        [222, 2.793, -7.81, 0.899],     # Final placement, near wall55/lamp193
    ]
]

generate_interaction(
    control_joints,
    control_points,
    text,
    number_frames,
    task_index,
    object_name,
    object_points
)
```


--------------------------------------------------------------------------------

Next speaker: Executor


>>>>>>>> EXECUTING CODE BLOCK 0 (inferred language is python)...
# Human and object interaction: Grasp and move the floorlamp to the target area near wall55 and lamp193, following collision-free trajectory

control_joints = ['pelvis']
control_points = [
    [
        [0, 5.043, -3.948, 0.89],       # Start: human initial position
        [11, 4.636, -3.783, 0.89],      # At floorlamp (grasp)
        [17, 4.568, -3.91, 0.89],       # Begin sidestep
        [43, 4.56, -4.535, 0.89],       # Curve around, avoiding table222
        [98, 3.668, -5.51, 0.89],       # Past table222, before sofa186
        [184, 3.668, -7.585, 0.89],     # Parallel to wall55, after sofa186
        [204, 3.193, -7.635, 0.89],     # Final approach to target area
        [222, 2.793, -7.81, 0.89],      # Final placement, near wall55/lamp193
    ]
]
text = (
    "The human grasps the floorlamp, lifts it, and carries it along a collision-free, curved path to a location near wall55 and lamp193. "
    "The path avoids sofa186, table222, wall55, and all other obstacles. The lamp is placed upright and safely adjacent to the target wall and lamp."
)
number_frames = 222
task_index = 2
object_name = ["floorlamp"]
object_points = [
    [
        [0, 4.636, -3.783, 0.899],      # Initial lamp position
        [11, 4.636, -3.783, 0.899],     # Grasp
        [17, 4.568, -3.91, 0.899],      # Begin sidestep
        [43, 4.56, -4.535, 0.899],      # Curve around, avoiding table222
        [98, 3.668, -5.51, 0.899],      # Past table222, before sofa186
        [184, 3.668, -7.585, 0.899],    # Parallel to wall55, after sofa186
        [204, 3.193, -7.635, 0.899],    # Final approach to target area
        [222, 2.793, -7.81, 0.899],     # Final placement, near wall55/lamp193
    ]
]

generate_interaction(
    control_joints,
    control_points,
    text,
    number_frames,
    task_index,
    object_name,
    object_points
)
{'lang': 'python', 'filename': None, 'use_docker': 'python:None'}
Executor (to Supervisor):

exitcode: 0 (execution succeeded)
Code output: 

***human***:
pelvis position: [3.28, -7.541, 0.978]
left hand position: [2.783, -7.72, 0.817]
right hand position: [2.741, -7.791, 0.835]
orientation: [-1.464, -1.279, -3.077]
***floorlamp***:
position: [2.793, -7.81, 0.899]
orientation: [0.001, -0.009, 1.957]
sizes: [0.252, 0.269, 1.761]
***beanbag237***:
position: [5.632, -3.66, 0.367]
orientation: [0.0, 0.002, 0.263]
sizes: [1.136, 1.087, 0.666]
***beanbag230***:
position: [5.552, -1.764, 0.383]
orientation: [0.0, 0.002, 0.251]
sizes: [1.21, 1.07, 0.728]
***lamp193***:
position: [2.774, -8.321, 0.741]
orientation: [0.001, 0.001, 0.697]
sizes: [0.351, 0.479, 0.611]
***wall55***:
position: [2.498, -7.187, 1.569]
orientation: [-0.0, 0.001, 0.075]
sizes: [0.054, 3.866, 3.061]
***table222***:
position: [4.309, -6.291, 0.257]
orientation: [-0.0, 0.002, 0.107]
sizes: [0.77, 1.289, 0.416]
... (remaining objects omitted for brevity)



--------------------------------------------------------------------------------

Next speaker: Planner

Planner (to Supervisor):

## Task Completion Verification

### Final State Overview

| Entity     | Position                | Orientation             | Notes                                   |
|------------|------------------------|-------------------------|-----------------------------------------|
| Human      | [3.28, -7.541, 0.978]  | [-1.464, -1.279, -3.077]| Hands close to lamp, pelvis near target |
| Floorlamp  | [2.793, -7.81, 0.899]  | [0.001, -0.009, 1.957]  | Upright, placed near wall55/lamp193     |
| wall55     | [2.498, -7.187, 1.569] | ...                     |                                         |
| lamp193    | [2.774, -8.321, 0.741] | ...                     |                                         |
| (others)   | ...                    | ...                     | Unchanged                               |

### Success Criteria Check

- **Object Position**: Floorlamp is at [2.793, -7.81, 0.899], which is within ±0.05m of the intended position adjacent to wall55 and lamp193.
- **Object Orientation**: [0.001, -0.009, 1.957] is within ±0.1 radians of upright and target pose.
- **Collision Avoidance**:  
  - Entire path was replanned for collision-free motion (A* smoothed).  
  - Final placement does not overlap envelopes of wall55, lamp193, or any other objects.
  - Human and floorlamp maintain >=0.2m clearance throughout movement and at final position.
- **Human Posture**: Hands and pelvis are positioned correctly for a natural placement of the object.

### Execution Summary (for record)

#### Step 1: [Approach Floorlamp]
- **Skipped** – Human was already within reach.

#### Step 2: [Manipulate Floorlamp]
- **Action**: Human grasped, lifted, and carried the floorlamp along a smooth, collision-free path to the designated target area.
- **Waypoints**: Executor's recommended, verified collision-free trajectory was used.
- **Number of Frames**: 222
- **Collision Verification**: All swept volumes and waypoints respected >=0.2m safety buffer throughout.

---

## Task Status: **Success**

- The floorlamp is now correctly and safely placed near wall55 and lamp193.
- All environmental and safety criteria are satisfied.
- No further steps are required.

**Task complete.**

--------------------------------------------------------------------------------

Next speaker: Engineer

Engineer (to Supervisor):

```python

task_completed()

```

--------------------------------------------------------------------------------

Next speaker: Executor


>>>>>>>> EXECUTING CODE BLOCK 0 (inferred language is python)...

task_completed()

{'lang': 'python', 'filename': None, 'use_docker': 'python:None'}
Executor (to Supervisor):

exitcode: 0 (execution succeeded)
Code output: 

Taskexec
\end{Verbatim}
\end{tcolorbox} 

\twocolumn

\end{document}